\documentclass{article}

\usepackage{arxiv}

\usepackage{amsmath, amsthm, amssymb, graphicx, hyperref, etoolbox, longtable, colonequals}

\makeatletter
\let\@nodottedtocline\@dottedtocline
\patchcmd{\@nodottedtocline}{\hbox{.}}{\hbox{}}{}{}
\patchcmd{\@nodottedtocline}{\normalcolor #5}{\normalcolor}{}{}
\newcommand*\l@sectionsubtitle{\@nodottedtocline{1}{0em}{1.5em}}
\makeatother

\usepackage{imakeidx}
\makeindex[title=Index,columns=3]
\indexsetup{noclearpage}

\usepackage{tikz}
\usetikzlibrary{calc,arrows.meta}

\theoremstyle{definition}

\newcommand\C{\mathbf{C}}

\newcommand\cofib\rightarrowtail

\newcommand\Hom{\textup{Hom}}

\newcommand\mdel[1]{}

\newcommand\Set{\textup{Set}}

\renewcommand\geq\geqslant
\renewcommand\leq\leqslant

\usepackage[utf8]{inputenc} 
\usepackage[T1]{fontenc}    
\usepackage{hyperref}       
\usepackage{url}            
\usepackage{booktabs}       
\usepackage{amsfonts}       
\usepackage{nicefrac}       
\usepackage{microtype}      
\usepackage{lipsum}		
\usepackage{graphicx}
\usepackage{natbib}
\usepackage{doi}
\usepackage{tikz-cd}
\usepackage{mathtools}

\usepackage{amsmath}
\usepackage{amssymb}
\usepackage{amsthm}
\usepackage[boxruled]{algorithm2e}

\newtheorem{theorem}{Theorem}
\newtheorem{definition}{Definition}
\newtheorem{lemma}{Lemma}
\newtheorem{example}{Example}

\newcommand{\CI}{\mathrel{\perp\mspace{-10mu}\perp}}

\title{A Layered Architecture for Universal Causality \thanks{Draft under revision. Comments welcome.} }

\author{ Sridhar Mahadevan \\
	Adobe Research and University of Massachusetts, Amherst\\
	\texttt{smahadev@adobe.com, mahadeva@umass.edu}
}

\hypersetup{
pdftitle={A template for the arxiv style},
pdfsubject={q-bio.NC, q-bio.QM},
pdfauthor={David S.~Hippocampus, Elias D.~Striatum},
pdfkeywords={First keyword, Second keyword, More},
}

\begin{document}
\maketitle

\begin{abstract}
We propose a layered hierarchical architecture called UCLA (Universal Causality Layered Architecture), which combines multiple levels of categorical abstraction for causal inference. At the top-most level, causal interventions are  modeled combinatorially using a simplicial category of ordinal numbers $\Delta$, whose objects are the ordered natural numbers $[n] = \{0, \ldots, n\}$, and whose morphisms are order-preserving injections and surjections. At the second layer, causal models are defined by a graph-type category consisting of a collection of objects, such as the entities in a relational database, and morphisms between objects can be viewed  as attributes relating entities. The non-random ``surgical" operations on causal structures, such as edge deletion, are captured using degeneracy and face operators from the simplicial layer above. The third categorical abstract layer corresponds to the data layer in causal inference, where each causal object is mapped into a set of instances, modeled using the category of sets and morphisms are functions between sets. The fourth homotopy layer comprises of additional structure imposed on the instance layer above, such as a topological space, a measurable space or a probability space, which enables evaluating causal models on datasets. Functors map between every pair of layers in UCLA. Each functor between layers is characterized by a universal arrow, which defines an isomorphism between every pair of categorical layers. These universal arrows define universal elements and representations through the Yoneda Lemma, and in turn lead to a new category of elements based on a construction introduced by Grothendieck. Causal inference between each pair of layers is defined as a lifting problem, a commutative diagram whose objects are categories, and whose morphisms are functors that are characterized as different types of fibrations. An interesting result is that monoidal categories can be shown to be special cases of simplicial objects under a particular type of fibration called a Grothendieck opfibration.  We illustrate the UCLA architecture using a range of examples, including integer-valued multisets that represent a non-graphical framework for conditional independence, and causal models based on graphs and string diagrams using symmetric monoidal categories. We define causal effect in terms of the homotopy colimit of the nerve of the category of elements.
\end{abstract}


\section{Introduction} 

\begin{figure}[h] 
\centering
\caption{UCLA is a layered architecture that defines Universal Causality. \label{ucla}}
\vskip 0.1in
\begin{minipage}{0.6\textwidth}
\vskip 0.1in
\includegraphics[scale=0.4]{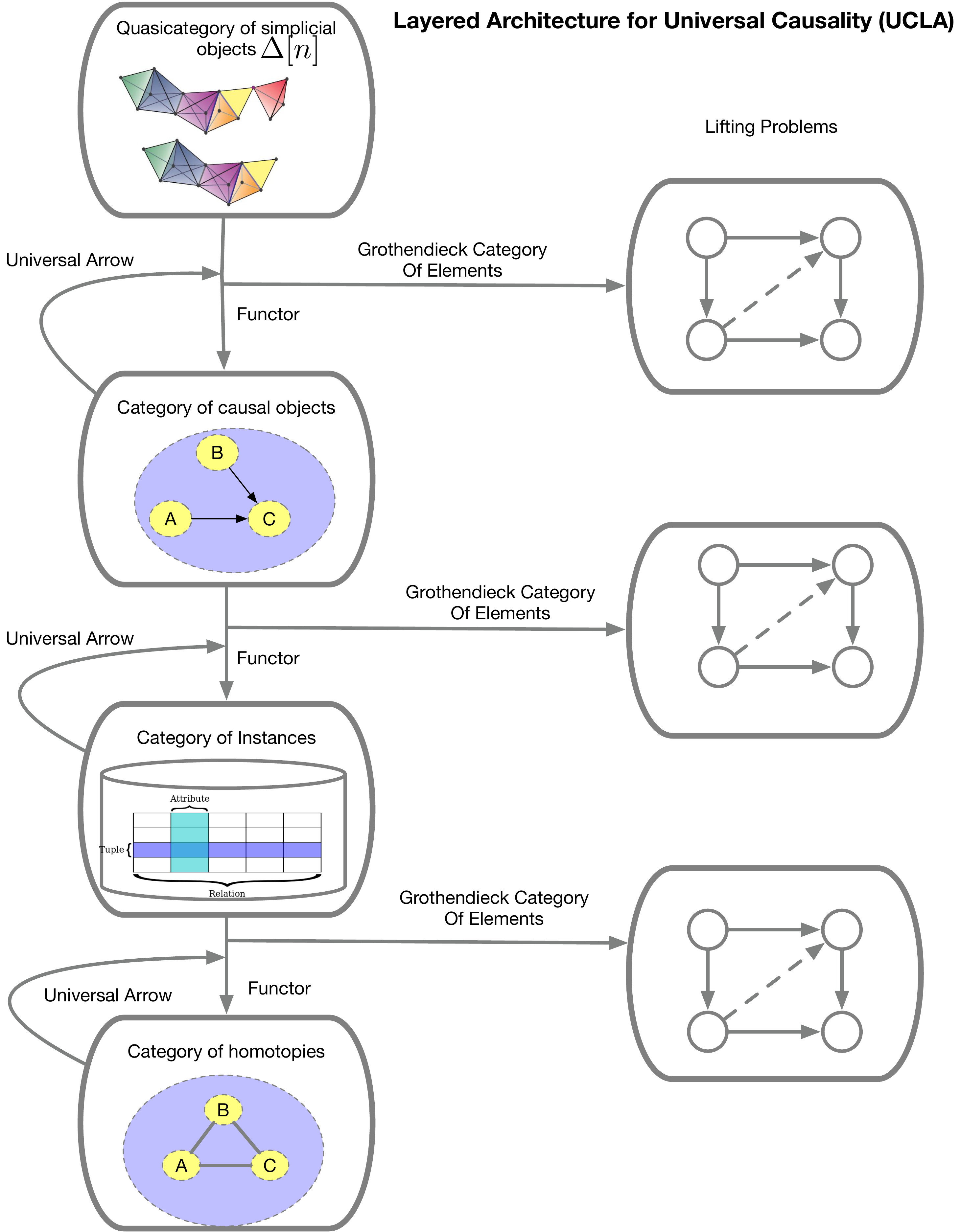}
\end{minipage}
\end{figure}

Universal Causality \cite{fong:ms,string-diagram-surgery,sm:udm,categoroids,sm:higher} is a framework for modeling causal inference using category theory, and includes a growing number of approaches, ranging from categorified representations of causal DAG models \cite{fong:ms,string-diagram-surgery} to higher-order categorical representations \cite{sm:higher}, categorical representations of conditional independence \cite{categoroids} to universal decision models combining causal inference with other types of decision making \cite{sm:udm}. In this paper, we propose a layered architecture that defines the framework called UCLA (Universal Causality Layered Architecture). This architecture is illustrated in Figure~\ref{ucla}.   Many variants are possible, as we will discuss in the paper. As functors compose with each other, it is also possible to consider ``collapsed" version of the UCLA hierarchy. 

\begin{table}[h]
 \caption{Each layer of UCLA represents a categorical abstraction of causal inference.}
 \centering
 \begin{small}
  \begin{tabular}{|c|c|c|c|} \hline 
{\bf Layer}  &  {\bf Objects} & {\bf Morphisms} & {\bf Description}  \\ \hline 
Simplicial & $[n] = \{0, 1, \ldots, n \}$ & $f= [m] \rightarrow [n]$ & Structure manipulations \\ \hline 
Graphical & Vertices $V$, Edges $E$ & $s, t: E \rightarrow V$ & Model interventions  \\ \hline
Instances & Sets & Functions on sets $f: S \rightarrow T$ & Causal Dataset  \\ \hline
Homotopy &  Topological Spaces & Continuous functions &  Find homotopic equivalences  \\ \hline 
\end{tabular}
\end{small}
\label{ucla-table}
\end{table}

\begin{figure}[h] 
\centering
\caption{Simplicial set structure of a causal model, along with three canonical causal structures. \label{causal-simplex}}
\vskip 0.1in
\begin{minipage}{0.9\textwidth}
\vskip 0.1in
\includegraphics[scale=0.22]{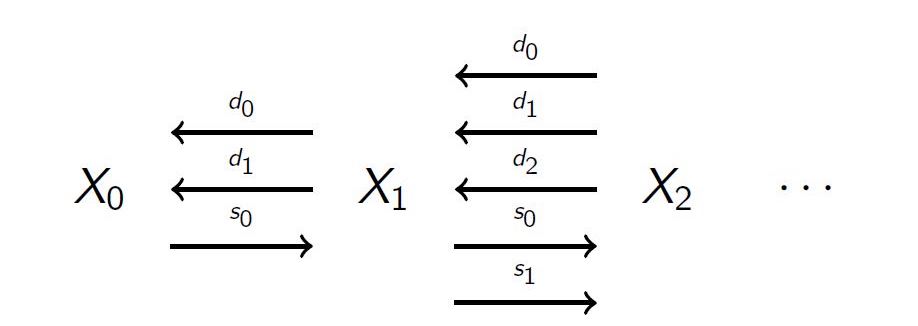}
\includegraphics[scale=0.2]{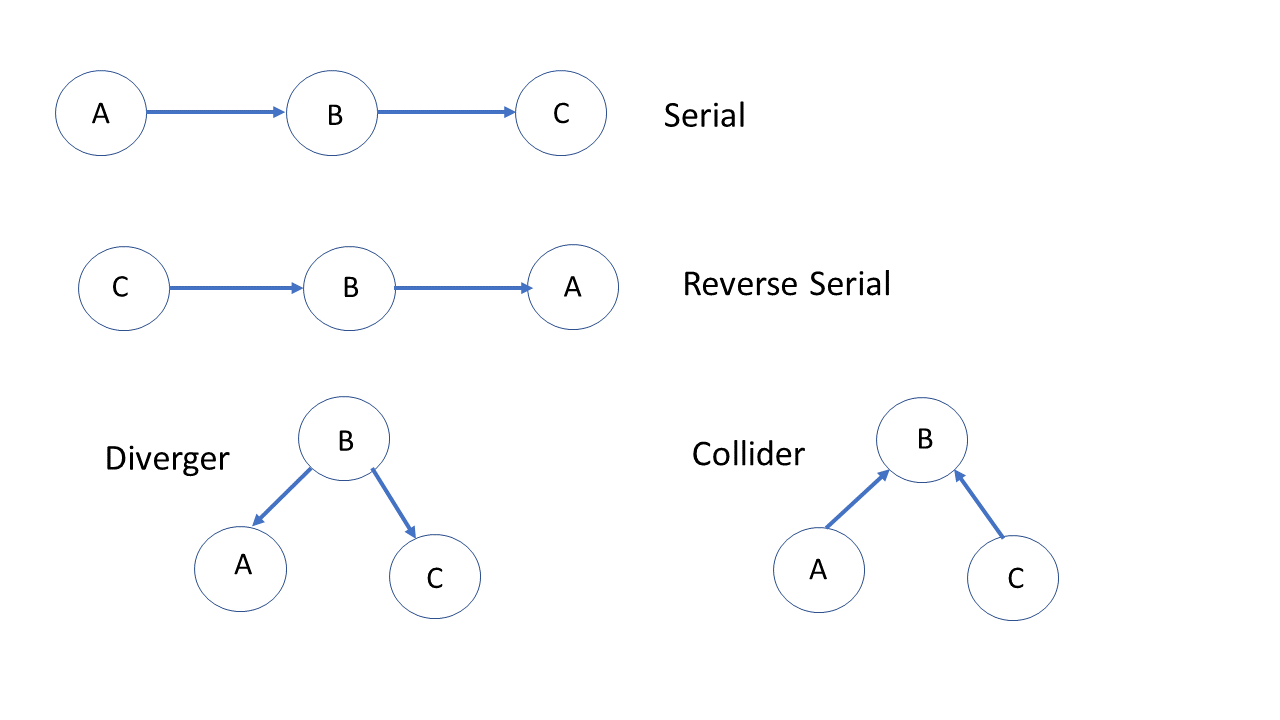}
\end{minipage}
\end{figure}

\begin{figure}[h] 
\centering
\caption{Universal arrows play a central role in the UCLA framework. In this example, the forgetful functor $U$ between {\bf Cat}, the category of all categories, and {\bf Graph}, the category of all (directed) graphs defines a universal arrow. The property asserts that every graph homomorphism problem defined by $\phi: G \rightarrow H$ uniquely factors through the universal graph homomorphism $u: G \rightarrow U(C)$ defined by the category $C$ defining the universal arrow property. In other words, the associated {\em extension} problem of ``completing" the triangle of graph homomorphisms in the category of {\bf Graph} can be uniquely solved by ``lifting" the associated category arrow $h: C \rightarrow D$. \label{universal-arrow}}
\vskip 0.1in
\begin{minipage}{0.6\textwidth}
\vskip 0.1in
\includegraphics[scale=0.4]{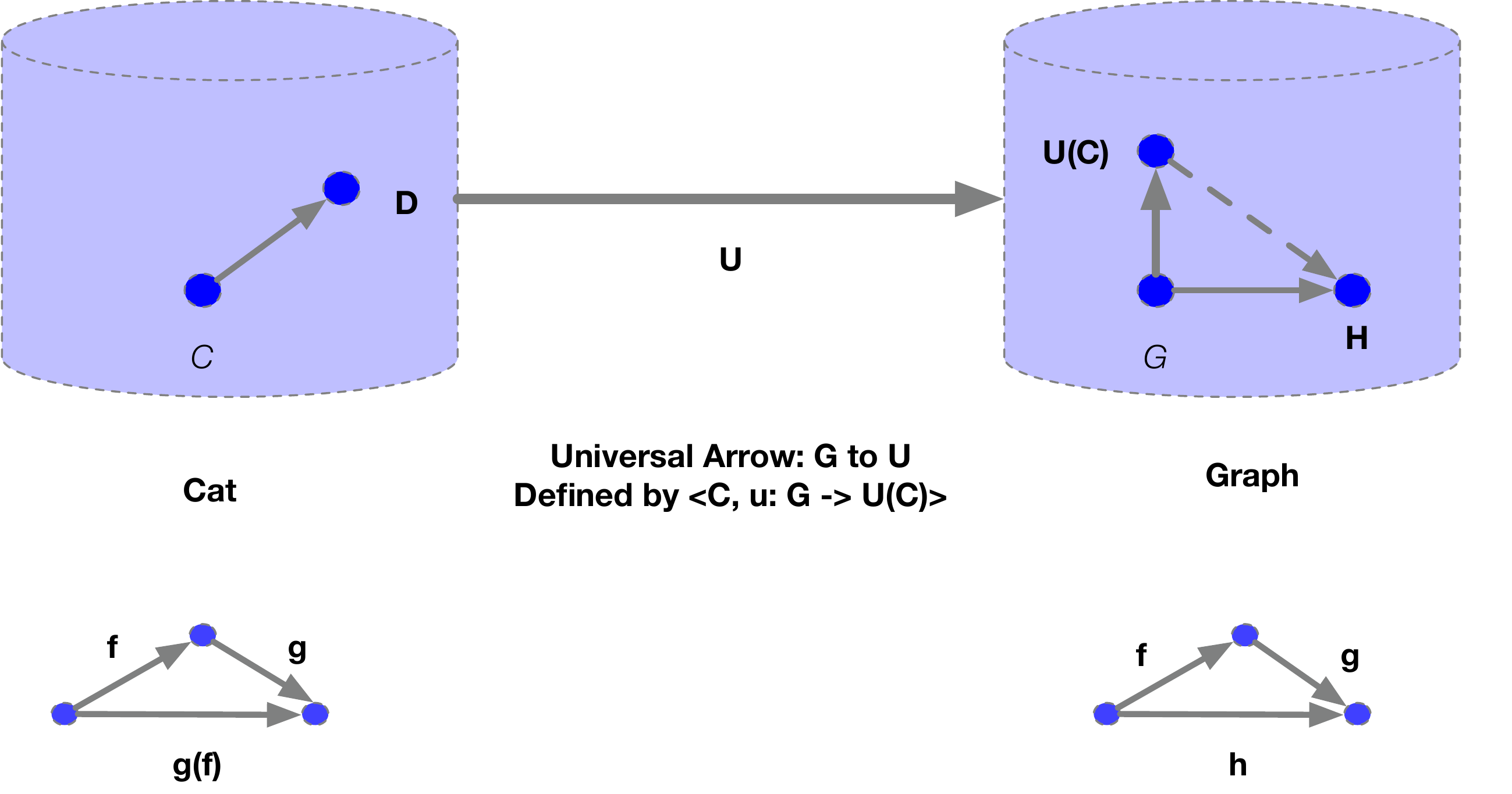}
\end{minipage}
\end{figure}

The UCLA architecture is built on the theoretical foundation of category theory \cite{maclane:71,maclane:sheaves,riehl2017category,JOYAL1996164}, a broad unified framework that has provided a unified mathematical language for over half a century.  Previous work has investigated using category theory to model causal inference. \citet{fong:ms} and \citet{string-diagram-surgery} both primarily focusing on reformulating Pearl's DAG model using symmetric monoidal categories to represent causal models \cite{string-diagram-surgery}. Our focus in this paper is elucidating the universal properties underlying causal inference in a representation-independent manner. We discuss the connections to \cite{fong:ms, string-diagram-surgery} at length later in the paper. At the heart of the UCLA hierarchy, as Figure~\ref{ucla} illustrates, is the principle of {\em universal arrows} \cite{maclane:71}, which we now explain with an example (see Figure~\ref{universal-arrow}). 

The fundamental property of {\em universal arrows} \cite{maclane:71} plays a central role in category theory and in the UCLA architecture. Figure~\ref{universal-arrow} explains this concept with an example, which also illustrates the connection between categories and graphs. For every (directed) graph $G$, there is a universal arrow from $G$ to the ``forgetful" functor $U$ mapping the category {\bf Cat} of all categories to {\bf Graph}, the category of all (directed) graphs, where for any category $C$, its associated graph is defined by $U(C)$. To understand this functor, simply consider a directed graph $U(C)$ as a category $C$ forgetting the rule for composition. That is, from the category $C$, which associates to each pair of composable arrows $f$ and $g$, the composed arrow $g \circ f$, we derive the underlying graph $U(G)$ simply by forgetting which edges correspond to elementary functions, such as $f$ or $g$, and which are composites. The universal arrow from a graph $G$ to the forgetful functor $U$  is defined as a pair $\langle G, u: G \rightarrow U(C) \rangle$, where $u$ is a a graph homomorphism. This arrow possesses the following {\em universal property}: for every other pair $\langle D, v: G \rightarrow H \rangle$, where $D$ is a category, and $v$ is an arbitrary graph homomorphism, there is a functor  $f': C \rightarrow D$, which is an arrow in the category {\bf Cat} of all categories, such that {\em every} graph homomorphism $\phi: G \rightarrow H$ uniquely factors through the universal graph homomorphism $u: G \rightarrow U(C)$  as the solution to the equation $\phi = U(f') \circ u$, where $U(f'): U(C) \rightarrow H$ (that is, $H = U(D)$).  Namely, the dotted arrow defines a graph homomorphism $U(f')$ that makes the triangle diagram ``commute", and the associated ``extension" problem of finding this new graph homomorphism $U(f')$ is solved by ``lifting" the associated category arrow $f': C \rightarrow D$. This property of universal arrows will, as we show in the paper, provide the conceptual underpinnings of the UCLA architecture, leading to the associated defining property of a universal causal representation through the Yoneda Lemma \cite{maclane:71}. 

Table~\ref{ucla-table} describes the composition of each layer. As Figure~\ref{ucla} illustrates, at the top layer of UCLA, we model causal inference  over {\em simplicial objects} \citep{may1992simplicial}, which is a combinatorial representation of a causal model. Intuitively, causal models are made of ``parts" (e.g., vertices and edges in a causal DAG \cite{pearl:causalitybook}, objects in a symmetric monoidal category defining a causal DAG \cite{string-diagram-surgery}, or elements of a semi-join lattice defining a conditional independence structure, such as an integer-valued multiset \cite{studeny2010probabilistic}). Causal interventions require ``surgery" of a causal model, taking parts away. We model this ``surgery" process formally by the use of simplicial objects. Formally, simplicial objects \cite{may1992simplicial}  are  contravariant functors $X: \Delta^{op} \rightarrow {\cal C}$ from the category of ordinal numbers, whose objects are $[n] = \{0,1, \ldots, n\}, n \geq 0$, and whose arrows are non-decreasing maps $f: [m] \rightarrow [n]$, into an underlying category ${\cal C}$. When the underlying category $C$ is the category of {\bf Sets}, we get simplicial sets.  We use the simplicial layer to capture ``surgery" of causal structures, including deletion of edges or more complex structures, which are represented using degeneracy and face operators. 

Figure~\ref{causal-simplex} illustrates the concept of causal simplicial structures. Here, $X$ denotes a causal structure represented as a category. $X[0]$ represents the ``objects" of the causal structure, defined formally as the contravariant functor $X[0]: [0] \rightarrow X$ from the simplicial category $\Delta$ to the causal category $X$. The arrows representing causal effects are defined as $X[1]: [1] \rightarrow X$. Note that since $[1]  = \{0, 1\}$ is a category by itself, it has one (non-identity) arrow $0 \rightarrow 1$ (as well as two identity arrows). The mapping of this arrow onto $X$ defines the ``edges" of the causal model. Similarly, $X_2$ represents ``triangles" of three objects, which can represent the canonical structures in a causal model, including serial nodes $A \rightarrow B \rightarrow C$, colliders $A \rightarrow B \leftarrow C$, and divergers $A \leftarrow B \rightarrow C$. Note that there is one edge from $X_0$ to $X_1$, labeled by $s_0$. This is a co-degeneracy operator from the simplicial layer that maps each object $A$ into an identity edge {\bf 1}$_A$. Similarly, there are two edges marked $d_0$ and $d_1$ from $X_1$ to $X_0$. These are co-face operators that map an edge to its source and target vertices correspondingly. Notice also that there are three edges from $X_2$ to $X_1$, marked $d_0$, $d_1$, and $d_2$. These are the ``faces" of each triangle, which take, for example, a collider structure $A \rightarrow B \leftarrow C$ 
and ``unpack" it in terms of its constituent edges. 

\begin{figure}[h] 
\centering
\caption{A causal model of climate change and Covid-19 lockdown. Universal causality interprets causal diagrams in terms of limits and co-limits of an indexing category of abstract diagrams. \label{lockdown}}
\vskip 0.1in
\begin{minipage}{0.8\textwidth}
\vskip 0.1in
\includegraphics[scale=0.4]{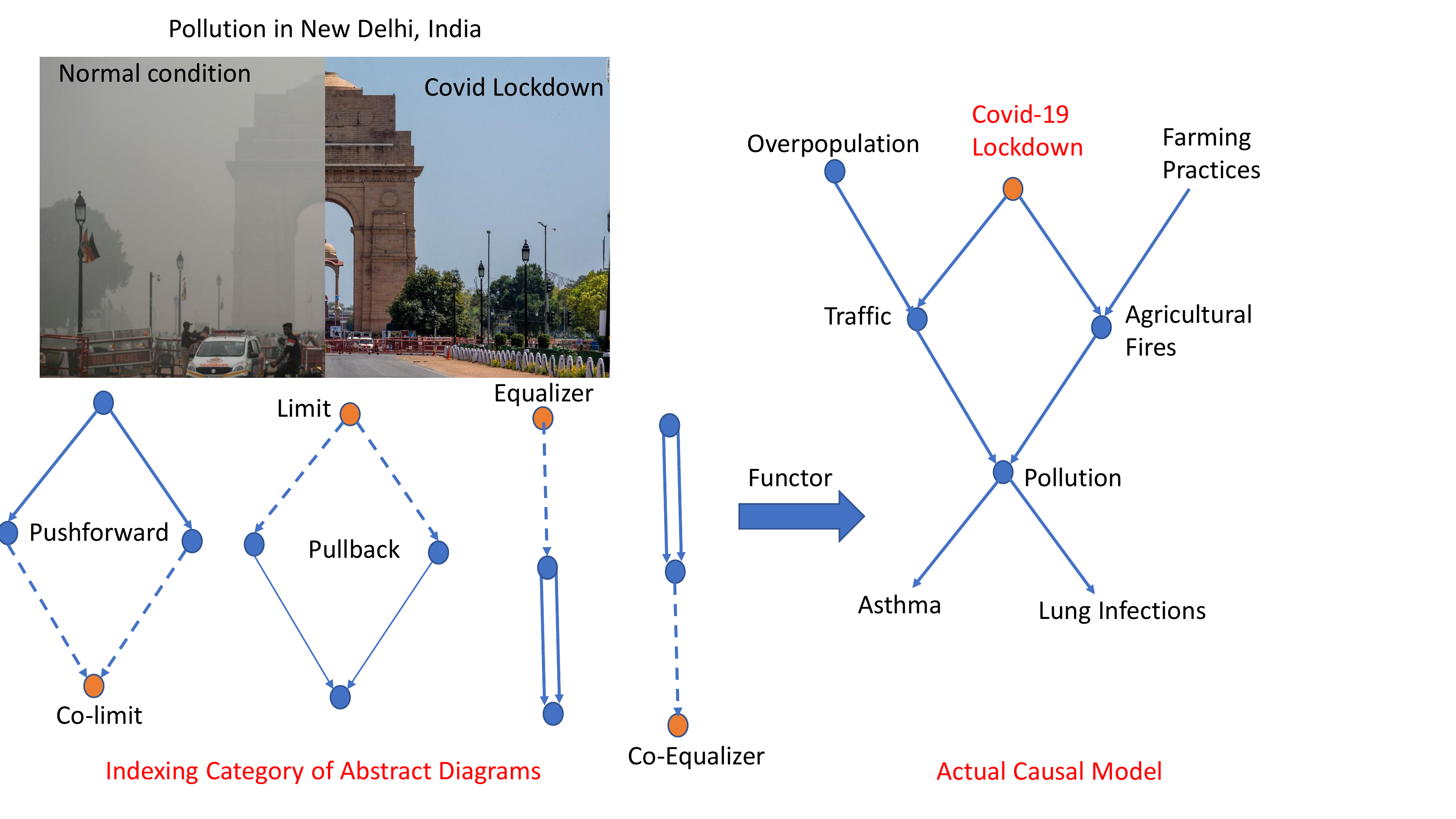}
\end{minipage}
\end{figure}

As shown in Figure~\ref{lockdown}, we can interpret causal DAG-like models in category theory in terms of functors mapping from some index category of diagrams into the actual causal model. Category theory defines {\em universal properties}, including limits, colimits, and equalizers, which helps clarify properties of causal diagrams. We will extensively build on the property of {\em universal arrows} \cite{maclane:71} in this paper, which serves as the conceptual underpinning how causal effects can be transmitted faithfully from one layer of the UCLA hierarchy to the next. 

Functors define the correspondence between each pair of categorical abstractions in Figure~\ref{ucla}. Thus, for example, the causal DAG model shown in layer 2 of the figure can be defined through a  contravariant functor $X: \Delta^{op} \rightarrow {\cal C}$, where ${\cal C}$ represents the category of causal models. The $0$th order simplicial objects defined as $X[0] = X_0$ represent the causal variables $A$, $B$, and $C$.  The $1$-simplices $X_1$ define directional causal relationships represented as $0$-simplices, or arrows in category ${\cal C}$, in this case, the two arrows between $A$ and $C$, and between $B$ and $C$. Associated with each such functor between two layers is a universal arrow \cite{maclane:71}, which specifies a universal property that ensures the functor faithfully represents the set of morphisms from the layer above to the layer below. Associated with the universal arrow is a construction defined by Grothendieck, which defines an associated category of elements $\int X$, whose objects are pairs $(n, i)$, where $n$ is an object in the category $\Delta$, and $i$ is an object in the category $X$, such that the morphism $f: [2] \rightarrow [0]$ in $\Delta$ is contravariantly mapped by the functor  to the edge between $A$ and $C$ in the causal model. 

As Figure~\ref{ucla} illustrates,  each pair of categorical layers is associated with a lifting problem that defines a language for expressing and answering queries between those levels. A lifting problem in category ${\cal C}$ is a commutative diagram $\sigma$ in ${\cal C}$ in which each composable pair of arrows defines path equivalences. \citet{lifting} shows a wide range of problems in topology and other areas can be represented as solutions to lifting problems. As a concrete application to causal discovery, recent work on the use of {\em separating sets} \cite{DBLP:conf/nips/KocaogluSB17} can be viewed as an instance of imposing a Kolmogorov $T_0$ topology on the variables in a causal model. The $T_0$ topology separation condition is an example of a property that can be stated in terms of a lifting problem. In simple terms, a category can be viewed as a directed graph, where certain paths can be defined as equivalent. So, in the diagram below, the path $nu \circ f = p \circ \mu$, where each arrow is labeled with a particular morphism. 
 
 \begin{center}
 \begin{tikzcd}
  A \arrow{d}{f} \arrow{r}{\mu}
    & X \arrow[red]{d}{p} \\
  B  \arrow[red]{r}[blue]{\nu}
&Y \end{tikzcd}
 \end{center} 

 A solution to a lifting problem in ${\cal C}$ is a morphism $h: B \rightarrow X$ in ${\cal C}$ satisfying $p \circ h = \nu$ and $h \circ f = \mu$ as indicated in the diagram below.
 
 \begin{center}
 \begin{tikzcd}
  A \arrow{d}{f} \arrow{r}{\mu}
    & X \arrow[red]{d}{p} \\
  B \arrow[ur,dashed, "h"] \arrow[red]{r}[blue]{\nu}
&Y \end{tikzcd}
 \end{center} 

As indicated in Figure~\ref{ucla}, each pair of layers in UCLA is accompanied by a lifting problem that is defined by the induced category of elements $\int F$, for some particular functor $F$. For example, the category of lifting problems between the causal model at the second layer from the top to the instance layer below it can be viewed as formalizing the process of inference in relational databases \cite{algebraic-databases,cdi,functorial-data-migration}. In particular, \citet{SPIVAK_2013} showed that database queries in languages like SQL can be formalized in terms of lifting problems. In this particular setting, the functor used maps a relational database model defined by a category ${\cal D}$ into a set of interlocking tables defined as a category of {\bf Sets}. Thus, relational databases can be viewed as set-valued functors $F: {\cal D} \rightarrow {\bf Sets}$. In this particular case, the associated lifting problem defined by the Grothendieck category of elements enables formalizing the process of answering SQL queries. The same construction, when applied to categories other than ${\bf Sets}$ gives us the ability to formulate causal inference over other categories, including topological spaces. 

 Many properties of Grothendieck's category of elements can be exploited (some of these are discussed in the context of relational database queries in \cite{SPIVAK_2013}), but for our application to causal inference, we are primarily interested in the associated class of lifting problems that define queries in a causal model. In particular, it can be shown that Grothendieck's category of elements formally defines a type of fibration called a {\em Grothendieck opfibration}. More specifically, if we consider any object $c \in {\cal C}$ in a category ${\cal C}$, and look at its fiber $\pi_\delta^{-1}(c)$, which forms a subcategory of the category of elements $I$, it can be shown that the projection of every morphism in the fiber into the category ${\cal C}$ is to the identity element ${\bf 1}_c$. \citet{richter2020categories} explains how monoidal categories themselves can be defined using simplicial objects using a Grothendieck opfibration. We refer the reader to these sources for additional details.  One implication of this property is that approaches to UC that build symmetric monoidal category representations of Bayesian networks \cite{fong:ms,string-diagram-surgery} can be formally shown to define special cases of the UCLA hierarchy. 
 
Finally, to explain the bottom-most layer in UCLA of homotopy categories, it is well known that many DAG models are not identifiable from observations alone \cite{pearl:causalitybook}.  Non-identifiability of DAG models, such as the serial model $A \rightarrow B \rightarrow C$, the diverging model $B \leftarrow A \rightarrow C$ and the reverse serial model $C \rightarrow B \rightarrow A$, occurs because given the parameterization of one of the models, the parameterization of the others can be derived using Bayes rule. For example, the serial model is parameterized in terms of the marginal probability $P(A)$, and the conditional probabilities $P(B |A)$ and $P(C | B)$. The diverger is parameterized using $P(B), P(A | B), P(C | B)$. Using Bayes rule, we have $P(A B) = P(A | B) P(B) = P(B | A) P(A)$, and similarly, $P(C B) = P(B | C) P(C) = P(C |B) P(B)$, we can derive the parameterization of the reverse serial model from the serial and diverger models. In contrast, for the converging model, $A \rightarrow B \leftarrow C$, its parameterization is in terms of $P(B | AC)$, which cannot be derived from the others using Bayes rule. 
 
More broadly, a fundamental premise underlying UC is that {\em presheaves} -- the set of morphisms {\bf Hom}$_{\cal C}(-, X)$ entering an object $X$ in a category ${\cal C}$ -- can serve as universal representers of causal inference.  Causal diagrams in UC are defined as functors, mapping from some abstract category of diagrams into a causal model, and use universal constructions, including the {\em pullback}, the {\em pushforward}, the {\em (co)equalizer}, and more generally, the {\em (co)limit} and the Kan extension. A standard construction in category theory shows that any set-valued contravariant functor can be represented as a co-limit of a diagram, which is defined through the category of elements \cite{maclane:sheaves}. The concept of representable presheaves comes from the Yoneda Lemma, which states that any presheaf is representable in the category {\bf Set}  by a fully faithful embedding.  From the Yoneda Lemma also follows straightforwardly the property that   the causal influence of any object $X$ on another object $Y$, denoted by {\bf Hom}$_{\cal C}(X, Y)$, can be defined  as a natural transformation between two representable presheaves. In most applications, presheaves carry significant additional structure, which can be modeled by an enriched category over monoidal categories, including rings, topological spaces, measurable spaces, and probability spaces.

Given this high level overview of UC and the UCLA architecture, in the reminder of the paper, we describe the various layers of the UCLA architecture in more detail, and explain the significance of the underlying conceptual ideas, including the use of the universal arrows and the category of elements from the Grothendieck construction. We will illustrate these concepts using a range of examples from previous literature. 

\section{Categories, Functors, and Universal Arrows}

As the relationship between categories and graphs is a close one, which has significant influence in causal inference, we define categories as essentially directed graphs equipped with a composition property \cite{borceux_1994}. The composition rule in category theory enables specifying which sets of paths should be considered equivalent, which manifests itself in the widespread use of commutative diagrams, as will be seen in this paper. Given a graph, we can define the ``free" category associated with it where we consider all possible paths between pairs of vertices (including self-loops) as the set of morphisms between them. In the reverse direction, given a category, we can define a ``forgetful" functor that extracts the underlying graph from the category, forgetting the composition rule. This process of going from one to the other embodies a fundamental underlying principle in category theory, called the {\em universal arrow} \cite{maclane:71}. 

\begin{definition}
\label{cat-defn}
A {\bf graph} ${\cal G}$ (sometimes referred to as a quiver) is a labeled directed multi-graph defined by a set $O$ of {\em objects}, a set $A$ of {\em arrows},  along with two morphisms $s: A \rightarrow O$ and $t: A \rightarrow O$ that specify the domain and co-domain of each arrow.  In this graph, we define the set of composable pairs of arrows by the set 

\[ A \times_O A = \{\langle g, f  \rangle | \ g, f \in A, \ \ s(g) = t(f) \} \]

A {\bf category } ${\cal C}$ is a graph ${\cal G}$ with two additional functions: ${\bf id}: O \rightarrow A$, mapping each object $c \in C$ to an arrow ${\bf id}_c$ and $\circ: A \times_O A \rightarrow A$, mapping each pair of composable morphisms $\langle f, g \rangle$ to their composition $g \circ f$. 
\end{definition}

It is worth emphasizing that no assumption is made here of the finiteness of a graph, either in terms of its associated objects (vertices) or arrows (edges). Indeed, it is entirely reasonable to define categories whose graphs contain an infinite number of edges. A simple example is the group $\mathbb{Z}$ of integers under addition, which can be represented as a single object, denoted $\{ \bullet \}$ and an infinite number of morphisms $f: \bullet \rightarrow \bullet$, each of which represents an integer, where composition of morphisms is defined by addition. In this example, all morphisms are invertible. In a general category with more than one object, a {\em groupoid} defines a category all of whose morphisms are invertible. 

As our paper focuses on the use of category theory to formalize causal inference, we interpret causal changes in terms of the concept of isomorphisms in category theory. We will elaborate this definition later in the paper. 

\begin{definition}
Two objects $X$ and $Y$ in a category ${\cal C}$ are deemed {\bf isomorphic}, or $X \cong Y$ if and only if there is an invertible morphism $f: X \rightarrow Y$, namely $f$ is both {\em left invertible} using a morphism $g: Y \rightarrow X$ so that $g \circ f = $ {\bf id}$_X$, and $f$ is {\em right invertible} using a morphism $h$ where $f \circ h = $ {\bf id}$_Y$. A {\bf causally isomorphic change} in a category is defined as a change of a causal object $Y$ into $\hat{Y}$ under an intervention that changes another object $X$ into $\hat{X}$ such that $\hat{Y} \cong Y$, that is, they are isomorphic. A {\bf causal non-isomorphic effect} is a change that leads to a non-isomorphic change where  $\hat{Y} \not \cong Y$. 
\end{definition}

In the category {\bf Sets}, two finite sets are considered isomorphic if they have the same number of elements, as it is then trivial to define an invertible pair of morphisms between them. In the category {\bf Vect}$_k$ of vector spaces over some field $k$, two objects (vector spaces) are isomorphic if there is a set of invertible linear transformations between them. As we will see below, the passage from a set to the ``free" vector space generated by elements of the set is another manifestation of the universal arrow property. 

 \begin{definition} 
A {\bf covariant functor} $F: {\cal C} \rightarrow {\cal D}$ from category ${\cal C}$ to category ${\cal D}$, and defined as the following: 
\begin{itemize} 
    \item An object $F X$ (sometimes written as $F(x)$) of the category ${\cal D}$ for each object $X$ in category ${\cal C}$.
    \item An  arrow  $F(f): F X \rightarrow F Y$ in category ${\cal D}$ for every arrow  $f: X \rightarrow Y$ in category ${\cal C}$. 
   \item The preservation of identity and composition: $F \ id_X = id_{F X}$ and $(F f) (F g) = F(g \circ f)$ for any composable arrows $f: X \rightarrow Y, g: Y \rightarrow Z$. 
\end{itemize}
\end{definition} 

\begin{definition} 
A {\bf contravariant functor} $F: {\cal C} \rightarrow {\cal D}$ from category ${\cal C}$ to category ${\cal D}$ is defined exactly like the covariant functor, except all the arrows are reversed. In the contravariant functor$F: C^{\mbox{op}} \rightarrow D$, every morphism $f: X \rightarrow Y$ is assigned the reverse morphism $F f: F Y \rightarrow F X$ in category ${\cal D}$. 
\end{definition} 

\begin{itemize} 
\item For every object $X$ in a category ${\cal C}$, there exists a covariant functor ${\cal C}(X, -): {\cal C} \rightarrow {\bf Set}$ that assigns to each object $Z$ in ${\cal C}$ the set of morphisms ${\cal C}(X,Z)$, and to each morphism $f: Y \rightarrow Z$, the pushforward mapping $f_*:{\cal C}(X,Y) \rightarrow {\cal C}(X, Z)$. 

\item For every object $X$ in a category ${\cal C}$, there exists a contravariant functor ${\cal C}(-, X): {\cal C}^{\mbox{op}} \rightarrow {\bf Set}$ that assigns to each object $Z$ in ${\cal C}$ the set of morphisms {\bf Hom}$_{\cal C}(X,Z)$, and to each morphism $f: Y \rightarrow Z$, the pullback mapping $f^*:$ {\bf Hom}$_{\cal C}(Z, X) \rightarrow {\cal C}(Y, X)$. Note how ``contravariance" implies the morphisms in the original category are reversed through the functorial mapping, whereas in covariance, the morphisms are not flipped.
\end{itemize} 

\subsection{Universal Arrows}

\begin{definition}
Given a functor $S: D \rightarrow C$ between two categories, and an object $c$ of category $C$, a {\bf universal arrow} from $c$ to $S$ is a pair $\langle r, u \rangle$, where $r$ is an object of $D$ and $u: c \rightarrow Sr$ is an arrow of $C$, such that the following universal property holds true: 

\begin{itemize} 
\item For every pair $\langle d, f \rangle$ with $d$ an object of $D$ and $f: c \rightarrow Sd$ an arrow of $C$, there is a unique arrow $f': r \rightarrow d$ of $D$ with $S f' \circ u = f$. 
\end{itemize}
\end{definition}

\begin{example}
Consider the example of universal arrows where {\bf Vect}$_K$ denotes the category of vector spaces over some field $K$, and its arrows correspond to linear transformations. Let $U$ denote the ``forgetful" functor $U: {\bf Vect}_K \rightarrow {\bf Set}$ that maps each vector space $V$ to its associated set of elements. Given any set $X$, we can define an associated vector space $V_X$ choosing the elements $x \in X$ as its basis vectors, and forming all formal linear combinations $\sum_i k_i x_i$, where $k_i \in K$ and $x_i \in X$. The mapping from each set element $x \in X$ into the corresponding vector in $V_X$ is an arrow $\lambda: X \rightarrow U(V_X)$. For any other vector space $W$, each function $\gamma: X \rightarrow U(W)$ can be transformed into a unique linear transformation $\delta: V_X \rightarrow W$ such that $U \delta \circ \lambda = \gamma$. Thus, here $\lambda$ acts as a universal arrow from the set $X$ to the forgetful functor $U$. 
\end{example}

\subsection{The Category of Fractions}

The problem of defining a category with a given subclass of invertible morphisms, called the category of fractions \citep{gabriel1967calculus}, is another concrete illustration of the close relationships between categories and graphs. It is also useful in the context of causal inference, as for example, in defining the Markov equivalence class of directed acyclic graphs (DAGs) as a category that is localized by considering all invertible arrows as isomorphisms. \citet{borceux_1994} has a detailed discussion of the ``calculus of fractions", namely how to define a category where a subclass of morphisms are to be treated as isomorphisms (e.g., the three models in Figure~\ref{imset} are equivalent in the sense that the morphisms are invertible). The formal definition is as follows: 

\begin{definition}
Consider a category ${\cal C}$ and a class $\Sigma$ of arrows of ${\cal C}$. The {\bf category of fractions} ${\cal C}(\Sigma^{-1})$ is said to exist when a category ${\cal C}(\Sigma^{-1})$ and a functor $\phi: {\cal C} \rightarrow {\cal C}(\Sigma^{-1})$ can be found with the following properties: 
\begin{enumerate}
    \item $\forall f, \phi(f)$ is an isomorphism. 
    \item If ${\cal D}$ is a category, and $F: {\cal C} \rightarrow {\cal D}$ is a functor such that for all morphisms $f \in \Sigma$, $F(f)$ is an isomorphism, then there exists a unique functor $G: {\cal C}(\Sigma^{-1}) \rightarrow {\cal D}$ such that $G \circ \phi = F$. 
\end{enumerate}
\end{definition}

A detailed construction of the category of fractions is given in \cite{borceux_1994}, which uses the underlying directed graph skeleton associated with the category. The characterization of the Markov equivalent class of ayclic directed graphs is an example of the abstract concept of category of fractions \citep{anderson-annals}. \footnote{Briefly, this condition states that two acyclic directed graphs are Markov equivalent if and only if they have the same skeleton and the same immoralities.} In our previous work \cite{sm:homotopy}, we explored  constructing homotopically invariant causal models over finite Alexandroff topological spaces, which can be seen as a special case of the UCLA framework.

\subsection{Lifting Problems}

The UCLA hierarchy is defined through a series of categorical abstractions of a causal model, ranging from a combinatorial model defined by a simplicial object down to a measure-theoretic or topological realization. Between each pair of layers, we can formulate a series of lifting problems \cite{lifting}.  Lifting problems provide elegant ways to define basic notions in a wide variety of areas in mathematics. For example, the notion of injective and surjective functions, the notion of separation in topology, and many other basic constructs can be formulated as solutions to lifting problems. Database queries in relational databases can be defined using lifting problems \cite{SPIVAK_2013}. Lifting problems define ways of decomposing structures into simpler pieces, and putting them back together again. 
 
 \begin{definition}
 Let ${\cal C}$ be a category. A {\bf lifting problem} in ${\cal C}$ is a commutative diagram $\sigma$ in ${\cal C}$. 
 
 \begin{center}
 \begin{tikzcd}
  A \arrow{d}{f} \arrow{r}{\mu}
    & X \arrow[red]{d}{p} \\
  B  \arrow[red]{r}[blue]{\nu}
&Y \end{tikzcd}
 \end{center} 
 \end{definition}
 
 \begin{definition}
 Let ${\cal C}$ be a category. A {\bf solution to a lifting problem} in ${\cal C}$ is a morphism $h: B \rightarrow X$ in ${\cal C}$ satisfying $p \circ h = \nu$ and $h \circ f = \mu$ as indicated in the diagram below.
 
 \begin{center}
 \begin{tikzcd}
  A \arrow{d}{f} \arrow{r}{\mu}
    & X \arrow[red]{d}{p} \\
  B \arrow[ur,dashed, "h"] \arrow[red]{r}[blue]{\nu}
&Y \end{tikzcd}
 \end{center} 
 \end{definition}
 
 \begin{definition}
 Let ${\cal C}$ be a category. If we are given two morphisms $f: A \rightarrow B$ and $p: X \rightarrow Y$ in ${\cal C}$, we say that $f$ has the {\bf left lifting property} with respect to $p$, or that p has the {\bf right lifting property} with respect to f if for every pair of morphisms $\mu: A \rightarrow X$ and $\nu: B \rightarrow Y$ satisfying the equations $p \circ \mu = \nu \circ f$, the associated lifting problem indicated in the diagram below.
 
 \begin{center}
 \begin{tikzcd}
  A \arrow{d}{f} \arrow{r}{\mu}
    & X \arrow[red]{d}{p} \\
  B \arrow[ur,dashed, "h"] \arrow[red]{r}[blue]{\nu}
&Y \end{tikzcd}
 \end{center} 
 
 admits a solution given by the map $h: B \rightarrow X$ satisfying $p \circ h = \nu$ and $h \circ f = \mu$. 
 \end{definition}
 
 \begin{example}
 Given the paradigmatic non-surjective morphism $f: \emptyset \rightarrow \{ \bullet \}$, any morphism p that has the right lifting property with respect to f is a {\bf surjective mapping}. 
 
 \begin{center}
 \begin{tikzcd}
  \emptyset \arrow{d}{f} \arrow{r}{\mu}
    & X \arrow[red]{d}{p} \\
  \{ \bullet \} \arrow[ur,dashed, "h"] \arrow[red]{r}[blue]{\nu}
&Y \end{tikzcd}
 \end{center} 

 \end{example}

  \begin{example}
 Given the paradigmatic non-injective morphism $f: \{ \bullet, \bullet \} \rightarrow \{ \bullet \}$, any morphism p that has the right lifting property with respect to f is an {\bf injective mapping}. 
 \begin{center}
 \begin{tikzcd}
  \{\bullet, \bullet \} \arrow{d}{f} \arrow{r}{\mu}
    & X \arrow[red]{d}{p} \\
  \{ \bullet \} \arrow[ur,dashed, "h"]  \arrow[red]{r}[blue]{\nu}
&Y \end{tikzcd}
 \end{center} 

 \end{example}

\section{Layers 1 and 2: Simplicial Objects and Causal Interventions} 

We now begin our discussion of the UCLA architecture shown earlier in Figure~\ref{ucla}, describing the top simplicial objects layer, and how it interacts with the causal category structure (layer 2). Abstractly, at the top layer, we want to define the categorical machinery to do ``diagram surgery", as in ``graph surgery" \cite{pearl:causalitybook} or ``string diagram surgery" \cite{string-diagram-surgery}. The structure of simplicial objects and sets helps us achieve this goal in a representation-independent manner, by using a category of ordinal numbers $[n] = \{0, \ldots, n \}$ with morphisms defined that respect the natural ordering, and using degeneracy and face operators to implement ``surgery" operations. 

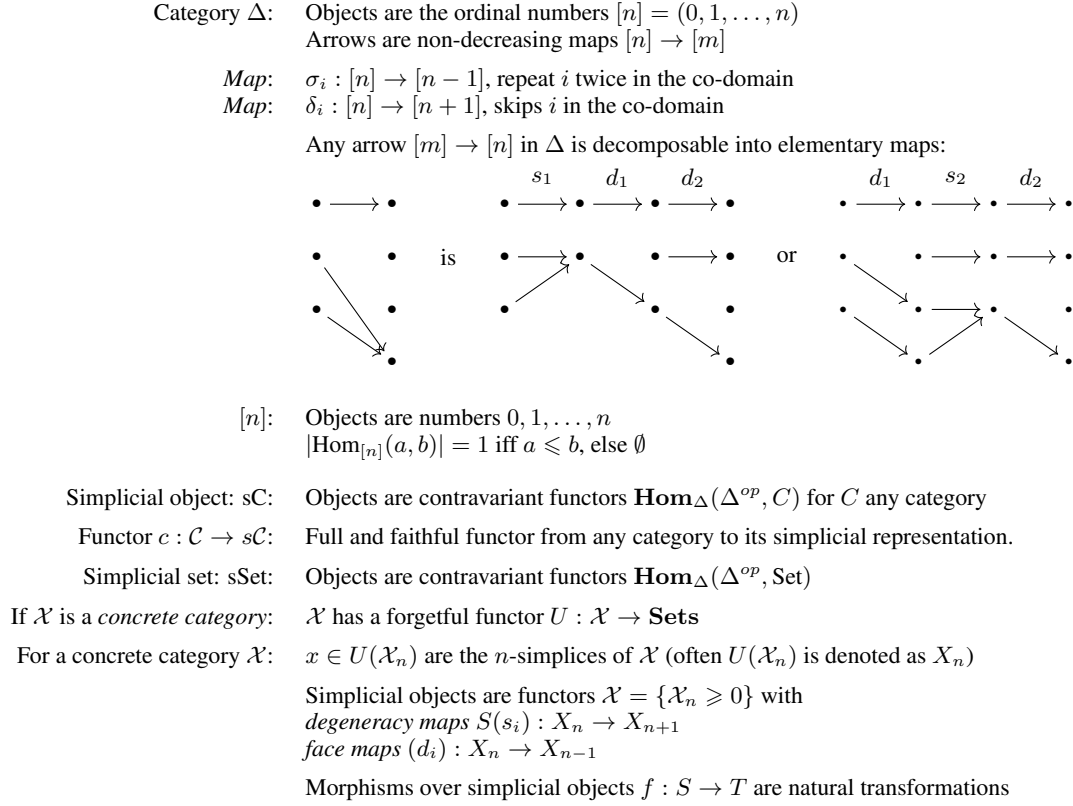
\begin{figure}[t] 
\centering
\begin{small}
\begin{minipage}{1.2\textwidth}
\begin{tabular}{r l l}
Category $\Delta$: & Objects are  the ordinal numbers $[n]=(0,1,\dots,n)$ \\ & \multicolumn{2}{l}{Arrows are non-decreasing  maps $[n]\to[m]$}\\[5pt]
\emph{Map}: & $\sigma_i: [n]\to [n-1]$, repeat $i$ twice in the co-domain& \\
\emph{Map}: &  $\delta_i:[n]\to [n+1]$, skips $i$ in the co-domain & \\[5pt]
& Any arrow $[m] \rightarrow [n]$ in $\Delta$ is decomposable into elementary maps: \\
& \multicolumn{2}{c}{
\begin{tikzpicture}[xscale=1,yscale=.7]
\foreach \x\y\namme in {0/1/1c, 0/2/1b, 0/3/1a, 1/0/2d, 1/1/2c, 1/2/2b, 1/3/2a}{
  \coordinate (\namme) at (\x,\y);
  \node[scale=.8] at (\x,\y) {$\bullet$};
}
\foreach \a\b in {1a/2a, 1b/2d, 1c/2d}{
  \draw[->,shorten >=5pt, shorten <=5pt] (\a)--(\b);
}
\begin{scope}[shift={(2.5,0)}]
\foreach \x\y\namme in {0/1/1c, 0/2/1b, 0/3/1a, 1/2/2b, 1/3/2a, 2/1/3c, 2/2/3b, 2/3/3a, 3/0/4d, 3/1/4c, 3/2/4b, 3/3/4a}{
  \coordinate (\namme) at (\x,\y);
  \node[scale=.8] at (\x,\y) {$\bullet$};
}
\foreach \a\b in {1a/2a, 1b/2b, 1c/2b, 2a/3a, 2b/3c, 3a/4a, 3b/4b, 3c/4d}{
  \draw[->,shorten >=5pt, shorten <=5pt] (\a)--(\b);
}
\node at (.5,3.5) {$s_1$};
\node at (1.5,3.5) {$d_1$};
\node at (2.5,3.5) {$d_2$};
\end{scope}
\begin{scope}[shift={(7,0)}]
\foreach \x\y\namme in {0/1/1c, 0/2/1b, 0/3/1a, 1/0/2d, 1/1/2c, 1/2/2b, 1/3/2a, 2/1/3c, 2/2/3b, 2/3/3a, 3/0/4d, 3/1/4c, 3/2/4b, 3/3/4a}{
  \coordinate (\namme) at (\x,\y);
  \node[scale=.6] at (\x,\y) {$\bullet$};
}
\foreach \a\b in {1a/2a, 1b/2c, 1c/2d, 2a/3a, 2b/3b, 2c/3c, 2d/3c, 3a/4a, 3b/4b, 3c/4d}{
  \draw[->,shorten >=5pt, shorten <=5pt] (\a)--(\b);
}
\node at (.5,3.5) {$d_1$};
\node at (1.5,3.5) {$s_2$};
\node at (2.5,3.5) {$d_2$};
\end{scope}
\node at (1.75,2) {is};
\node at (6.25,2) {or};
\end{tikzpicture}
}\\[10pt]
$[n]$: & Objects are numbers $0,1,\dots,n$ \\ & $|\Hom_{[n]}(a,b)| = 1$ iff $a\leqslant b$, else $\emptyset$\\[10pt]
Simplicial object:  sC: & Objects are contravariant functors ${\bf Hom}_{\Delta}(\Delta^{op},C)$ for $C$ any category \\ [5pt]
Functor $c: {\cal C} \rightarrow s{\cal C}$: & Full and faithful functor from any category to its simplicial representation. \\ [5pt]
Simplicial set: sSet: & Objects are contravariant functors ${\bf Hom}_{\Delta}(\Delta^{op},\Set)$ \\[5pt]
If ${\cal X}$ is a {\em concrete category}: & ${\cal X}$ has a forgetful functor $U: {\cal X} \rightarrow {\bf Sets}$ \\ [5pt]
For a concrete category ${\cal X}$: & $x \in U({\cal X}_n)$ are the $n$-simplices of ${\cal X}$ (often $U({\cal X}_n)$ is denoted as $X_n$) \\ [5pt]
& \multicolumn{2}{l}{Simplicial objects are functors ${\cal X} = \{{\cal X}_{n} \geqslant 0 \}$ with} \\
& \emph{degeneracy maps} $S(s_i):X_n\to X_{n+1}$ & \\
& \emph{face maps} $(d_i):X_n\to X_{n-1}$ & \\[5pt]
& \multicolumn{2}{l}{Morphisms over simplicial objects $f:S\to T$ are natural transformations} \\[10pt]
\end{tabular}

\vskip -0.5in
\end{minipage}
\end{small}
\caption{Simplicial objects are a combinatorial representation for manipulating causal categories.} \label{simpobj}
 \end{figure}

Simplicial objects have long been a foundation for algebraic topology \citep{may1999concise,may1992simplicial}, and  more recently in  infinite category theory \citep{weakkan,quasicats,kerodon}. The category $\Delta$ has non-empty ordinals $[n] = \{0, 1, \ldots, n]$ as objects, and order-preserving maps $[m] \rightarrow [n]$ as arrows. An important property in $\Delta$ is that any many-to-many mapping is decomposable as a composition of an injective and a surjective mapping,  each of which is decomposable into a sequence of elementary injections $\delta_i: [n] \rightarrow [n+1]$, called {\em coface} mappings, which omits $i \in [n]$, and a sequence of elementary surjections $\sigma_i: [n] \rightarrow [n-1]$, called {\em co-degeneracy} mappings, which repeats $i \in [n]$. The fundamental simplex $\Delta([n])$ is the presheaf of all morphisms into $[n$, that is, the representable functor $\Delta(-, [n])$.  The Yoneda Lemma \cite{maclane:71} assures us that an $n$-simplex $x \in X_n$ can be identified with the corresponding map $\Delta[n] \rightarrow X$. Every morphism $f: [n] \rightarrow [m]$ in $\Delta$ is functorially mapped to the map $\Delta[m] \rightarrow \Delta[n]$ in ${\cal S}$. 
See Figure~\ref{simpobj}. 

Any morphism in the category $\Delta$ can be defined as a sequence of {\em co-degeneracy} and {\em co-face} operators, where the co-face operator $\delta_i: [n-1] \rightarrow [n], 0 \leq i \leq n$ is defined as: 

\[ 
\delta_i (j)  =
\left\{
	\begin{array}{ll}
		j,  & \mbox{for } \ 0 \leq j \leq i-1 \\
		j+1 & \mbox{for } \  i \leq j \leq n-1 
	\end{array}
\right. \] 

Analogously, the co-degeneracy operator $\sigma_j: [n+1] \rightarrow [n]$ is defined as 

\[ 
\sigma_j (k)  =
\left\{
	\begin{array}{ll}
		j,  & \mbox{for } \ 0 \leq k \leq j \\
		k-1 & \mbox{for } \  j < k \leq n+1 
	\end{array}
\right. \] 

Note that under the contravariant mappings, co-face mappings turn into face mappings, and co-degeneracy mappings turn into degeneracy mappings. That is, for any simplicial object (or set) $X_n$, we have $X(\delta_i) \coloneqq d_i: X_n \rightarrow X_{n-1}$, and likewise, $X(\sigma_j) \coloneqq s_j: X_{n-1} \rightarrow X_n$. 

The compositions of these arrows define certain well-known properties \citep{may1992simplicial,richter2020categories}: 

\begin{eqnarray*}
    \delta_j \circ \delta_i &=& \delta_i \circ \delta_{j-1}, \ \ i < j \\
    \sigma_j \circ \sigma_i &=& \sigma_i \circ \sigma_{j+1}, \ \ i \leq j \\ 
    \sigma_j \circ \delta_i (j)  &=&
\left\{
	\begin{array}{ll}
		\sigma_i \circ \sigma_{j+1},  & \mbox{for } \ i < j \\
		1_{[n]} & \mbox{for } \  i = j, j+1 \\ 
		\sigma_{i-1} \circ \sigma_j, \mbox{for} \ i > j + 1
	\end{array}
\right.
\end{eqnarray*}

\begin{example}
The ``vertices" of a simplicial object ${\cal C}_n$ are the objects in  ${\cal C}$, and the ``edges" of ${\cal C}$ are its arrows $f: X \rightarrow Y$, where $X$ and $Y$ are objects in ${\cal C}$. Given any such arrow, the degeneracy operators $d_0 f = Y$ and $d_1 f = X$ recover the source and target of each arrow. Also, given an object $X$ of category ${\cal C}$, we can regard the face operator $s_0 X$ as its identity morphism ${\bf 1}_X: X \rightarrow X$. 
\end{example}

\begin{example} 
Given a category ${\cal C}$, we can identify an $n$-simplex $\sigma$ of a simplicial set ${\cal C}_n$ with the sequence: 

\[ \sigma = C_o \xrightarrow[]{f_1} C_1 \xrightarrow[]{f_2} \ldots \xrightarrow[]{f_n} C_n \] 

the face operator $d_0$ applied to $\sigma$ yields the sequence 

\[ d_0 \sigma = C_1 \xrightarrow[]{f_2} C_2 \xrightarrow[]{f_3} \ldots \xrightarrow[]{f_n} C_n \] 

where the object $C_0$ is ``deleted" along with the morphism $f_0$ leaving it.

\end{example} 

\begin{example} 
Given a category ${\cal C}$, and an $n$-simplex $\sigma$ of the simplicial set ${\cal C}_n$, the face operator $d_n$ applied to $\sigma$ yields the sequence 

\[ d_n \sigma = C_0 \xrightarrow[]{f_1} C_1 \xrightarrow[]{f_2} \ldots \xrightarrow[]{f_{n-1}} C_{n-1} \] 

where the object $C_n$ is ``deleted" along with the morphism $f_n$ entering it.  Note this process can be used to implement ``surgery" of a causal model, such as a causal DAG \cite{pearl:causalitybook}, or a symmetric monoidal category \cite{string-diagram-surgery}. 

\end{example} 

\begin{example} 
Given a category  ${\cal C}$, and an $n$-simplex $\sigma$ of the simplicial set ${\cal C}_n$
the face operator $d_i, 0 < i < n$ applied to $\sigma$ yields the sequence 

\[ d_i \sigma = C_0 \xrightarrow[]{f_1} C_1 \xrightarrow[]{f_2} \ldots C_{i-1} \xrightarrow[]{f_{i+1} \circ f_i} C_{i+1} \ldots \xrightarrow[]{f_{n}} C_{n} \] 

where the object $C_i$ is ``deleted" and the morphisms $f_i$ is composed with morphism $f_{i+1}$.  Note that this process can be abstractly viewed as intervening on object $C_i$ by choosing a specific value for it (which essentially ``freezes" the morphism $f_i$ entering object $C_i$ to a constant value). 

\end{example} 

\begin{example} 
Given a category ${\cal C}$, and an $n$-simplex $\sigma$ of the simplicial set ${\cal C}_n$, 
the degeneracy operator $s_i, 0 \leq i \leq n$ applied to $\sigma$ yields the sequence 

\[ s_i \sigma = C_0 \xrightarrow[]{f_1} C_1 \xrightarrow[]{f_2} \ldots C_{i} \xrightarrow[]{{\bf 1}_{C_i}} C_{i} \xrightarrow[]{f_{i+1}} C_{i+1}\ldots \xrightarrow[]{f_{n}} C_{n} \] 

where the object $C_i$ is ``repeated" by inserting its identity morphism ${\bf 1}_{C_i}$. 

\end{example} 

\begin{definition} 
Given a category ${\cal C}$, and an $n$-simplex $\sigma$ of the simplicial set ${\cal C}_n$, 
$\sigma$ is a {\bf degenerate} simplex if some $f_i$ in $\sigma$  is an identity morphism, in which case $C_i$ and $C_{i+1}$ are equal. 
\end{definition} 

\subsection{Simplicial Subsets and Horns}

We now describe more complex ways of extracting parts of causal structures using simplicial subsets and horns. These structures will play a key role in defining suitable lifting problems. 
 
 \begin{definition}
 The {\bf standard simplex} $\Delta^n$ is the simplicial set defined by the construction 
 
 \[ ([m] \in \Delta) \mapsto {\bf Hom}_\Delta([m], [n]) \] 
 
 By convention, $\Delta^{-1} \coloneqq \emptyset$. The standard $0$-simplex $\Delta^0$ maps each $[n] \in \Delta^{op}$ to the single element set $\{ \bullet \}$. 
 \end{definition}
 
 \begin{definition}
 Let $S_\bullet$ denote a simplicial set. If for every integer $n \geq 0$, we are given a subset $T_n \subseteq S_n$, such that the face and degeneracy maps 
 
 \[ d_i: S_n \rightarrow S_{n-1} \ \ \ \ s_i: S_n \rightarrow S_{n+1} \] 
 
 applied to $T_n$ result in 
 
 \[ d_i: T_n \rightarrow T_{n-1} \ \ \ \ s_i: T_n \rightarrow T_{n+1} \] 
 
 then the collection $\{ T_n \}_{n \geq 0}$ defines a {\bf simplicial subset} $T_\bullet \subseteq S_\bullet$
 \end{definition}
 
 \begin{definition}
 The {\bf boundary} is a simplicial set $(\partial \Delta^n): \Delta^{op} \rightarrow {\bf Set}$ defined as
 
 \[ (\partial \Delta^n)([m]) = \{ \alpha \in {\bf Hom}_\Delta([m], [n]): \alpha \ \mbox{is not surjective} \} \]
 \end{definition}
 
 Note that the boundary $\partial \Delta^n$ is a simplicial subset of the standard $n$-simplex $\Delta^n$. 
 
 \begin{definition}
 The {\bf Horn} $\Lambda^n_i: \Delta^{op} \rightarrow {\bf Set}$ is defined as
 
 \[ (\Lambda^n_i)([m]) = \{ \alpha \in {\bf Hom}_\Delta([m],[n]): [n] \not \subseteq \alpha([m]) \cup \{i \} \} \] 
 \end{definition}
 
 Intuitively, the Horn $\Lambda^n_i$ can be viewed as the simplicial subset that results from removing the interior of the $n$-simplex $\Delta^n$ together with the face opposite its $i$th vertex.

\subsection{Example: Causal Intervention and Horn Filling of Simplicial Complexes}

\begin{figure}[h] 
\centering
\vskip 0.1in
\begin{minipage}{0.5\textwidth}
\includegraphics[scale=0.4]{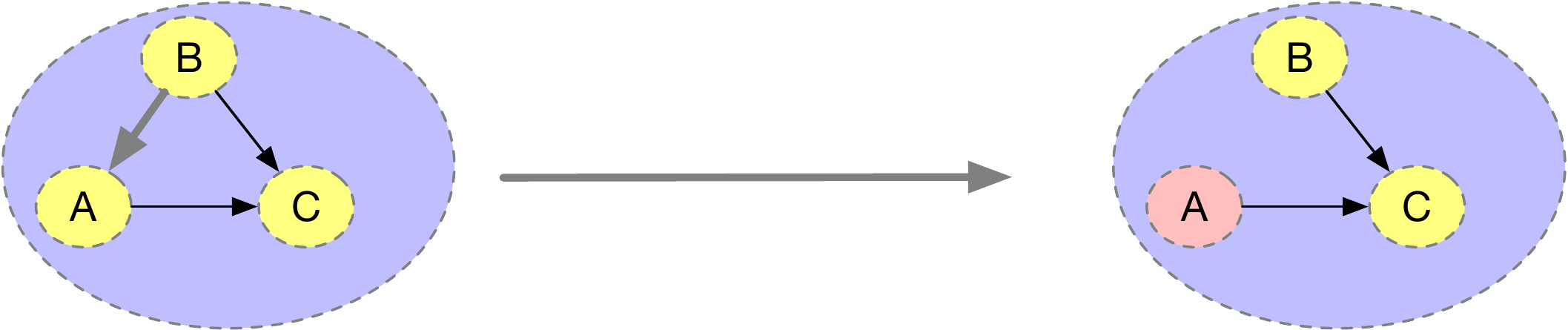}
\end{minipage}
\caption{Causal interventions can be related to horn filling of a simplicial complex. \label{horn-simplex} }
 \end{figure}

 Let us illustrate this abstract discussion above by instantiating it in the context of causal inference. Figure~\ref{horn-simplex} instantiates the abstract discussion above in terms of an example from causal inference. We are given a simple 3 variable DAG, on which we desire to explore the causal effect of variable $A$ on $C$. Using Pearl's backdoor criterion, we can intervene on variable $A$ by freezing its value $do(A=1)$, for example, which will eliminate the dependence of $A$ on $B$. Consider now the lifting problem where we want to know if there is a completion of this simplicial subset $\Lambda^2_2$, which is a ``outer horn" 

We can view the causal intervention problem in the more abstract setting of a class of lifting problem, shown  with the following diagrams. Consider the problem of composing $1$-dimensional simplices  to form a $2$-dimensional simplicial object. Each simplicial subset of an $n$-simplex induces a  a {\em horn} $\Lambda^n_k$, where  $ 0 \leq k \leq n$. Intuitively, a horn is a subset of a simplicial object that results from removing the interior of the $n$-simplex and the face opposite the $i$th vertex. Consider the three horns defined below. The dashed arrow  $\dashrightarrow$ indicates edges of the $2$-simplex $\Delta^2$ not contained in the horns. 

\begin{center}
 \begin{tikzcd}[column sep=small]
& \{0\}  \arrow[dl] \arrow[dr] & \\
  \{1 \} \arrow[rr, dashed] &                         & \{ 2 \} 
\end{tikzcd} \hskip 0.5 in 
 \begin{tikzcd}[column sep=small]
& \{0\}  \arrow[dl] \arrow[dr, dashed] & \\
  \{1 \} \arrow{rr} &                         & \{ 2 \} 
\end{tikzcd} \hskip 0.5in 
 \begin{tikzcd}[column sep=small]
& \{0\}  \arrow[dl, dashed] \arrow[dr] & \\
  \{1 \} \arrow{rr} &                         & \{ 2 \} 
\end{tikzcd}
\end{center}

The inner horn $\Lambda^2_1$ is the middle diagram above, and admits an easy solution to the ``horn filling" problem of composing the simplicial subsets. The two outer horns on either end pose a more difficult challenge. A considerable elaboration of the theoretical machinery in category theory is required to describe the various solutions proposed, which led to different ways of defining higher-order category theory \citep{weakkan,quasicats,kerodon}. Note that in our causal intervention example above shown in Figure~\ref{horn-simplex}, the horn filling problem is at the rightmost (where we map $A \rightarrow \{1 \}, B \rightarrow \{0 \}, C \rightarrow \{ 2 \}$. We know that this lifting problem indeed has a solution because the completion problem can be solved as the simplicial subset was created by intervention on node $A$. To elaborate on how we can solve such a lifting problem, we will postpone this discussion to the next section, where we show that causal interventions can be modeled as lifting problems in the category of elements (which will enable us to ``bind" objects such as $A$ to constants, such as $A=1$). For the sake of completeness, we include a more detailed discussion of the horn filling problem, and its various solutions in higher-order category theory. 
 
\subsection{Homotopy of Simplicial Objects}

We will discuss homotopy of causal structures in more detail below, for now we briefly discuss a few notions that will be useful in this section. 

 \begin{definition}
 Let $C$ and $C'$ be a pair of objects in a category ${\cal C}$. We say $C$ is {\bf a retract} of $C'$ if there exists maps $i: C \rightarrow C'$ and $r: C' \rightarrow C$ such that $r \circ i = \mbox{id}_{\cal C}$. 
 \end{definition}
 
 \begin{definition}
 Let ${\cal C}$ be a category. We say a morphism $f: C \rightarrow D$ is a {\bf retract of another morphism} $f': C \rightarrow D$ if it is a retract of $f'$ when viewed as an object of the functor category ${\bf Hom}([1], {\cal C})$. A collection of morphisms $T$ of ${\cal C}$ is {\bf closed under retracts} if for every pair of morphisms $f, f'$ of ${\cal C}$, if $f$ is a retract of $f'$, and $f'$  is in $T$, then $f$ is also in $T$. 
 \end{definition}

 \begin{definition}
  Let X and Y be simplicial sets, and suppose we are given a pair of morphisms $f_0, f_1: X \rightarrow Y$. A {\bf homotopy} from $f_0$ to $f_1$ is a morphism $h: \Delta^1 \times X \rightarrow Y$ satisfying $f_0 = h |_{{0} \times X}$ and $f_1 = h_{ 1 \times X}$. 
 \end{definition}
 
 This notion of homotopy generalizes the notion of homotopy in topology, which defines why an object like a coffee cup is topologically homotopic to a doughnut (they have the same number of ``holes"). 
 
 \subsection{Fibrations and Kan Complexes} 
 
 \begin{definition}
 Let $f: X \rightarrow S$ be a morphism of simplicial sets. We say $f$ is a {\bf Kan fibration} if, for each $n > 0$, and each $0 \leq i \leq n$, every lifting problem 
 
 \begin{center}
 \begin{tikzcd}
  \Lambda^n_i \arrow{d}{} \arrow{r}{\sigma_0}
    & X \arrow[red]{d}{f} \\
  \Delta^n \arrow[ur,dashed, "\sigma"] \arrow[red]{r}[blue]{\bar{\sigma}}
&S \end{tikzcd}
 \end{center} 
 
 admits a solution. More precisely, for every map of simplicial sets $\sigma_0: \Lambda^n_i \rightarrow X$ and every $n$-simplex $\bar{\sigma}: \Delta^n \rightarrow S$ extending $f \circ \sigma_0$, we can extend $\sigma_0$ to an $n$-simplex $\sigma: \Delta^n \rightarrow X$ satisfying $f \circ \sigma = \bar{\sigma}$. 
 \end{definition}
 
 \begin{example}
Given a simplicial set $X$, then a projection map $X \rightarrow \Delta^0$ that is a Kan fibration is called a {\bf Kan complex}. 
\end{example} 

\begin{example}
Any isomorphism between simplicial sets is a Kan fibration. 
\end{example}

\begin{example}
The collection of Kan fibrations is closed under retracts. 
\end{example}

\subsection{Higher-order Categories} 

We now formally introduce higher-order categories, building on the framework proposed in a number of formalisms \citep{weakkan,quasicats,kerodon}. 

\begin{definition}\citep{kerodon}
\label{ic} 
An $\infty$-category is a simplicial object $S_\bullet$ which satisfies the following condition: 

\begin{itemize} 
\item For $0 < i < n$, every map of simplicial sets $\sigma_0: \Lambda^n_i \rightarrow S_\bullet$ can be extended to a map $\sigma: \Delta^n \rightarrow S_i$. 
\end{itemize} 
\end{definition}

This definition emerges out of a common generalization of two other conditions on a simplicial set $S_i$: 

\begin{enumerate} 
\item {\bf Property K}: For $n > 0$ and $0 \leq i \leq n$, every map of simplicial sets $\sigma_0: \Lambda^n_i \rightarrow S_\bullet$ can be extended to a map $\sigma: \Delta^n \rightarrow S_i$. 

\item {bf Property C}:  for $0 < 1 < n$, every map of simplicial sets $\sigma_0: \Lambda^n_i \rightarrow S_i$ can be extended uniquely to a map $\sigma: \Delta^n \rightarrow S_i$. 
\end{enumerate} 

Simplicial objects that satisfy property K were defined above to be Kan complexes. Simplicial objects that satisfy property C above can be identified with the nerve of a category, which yields a full and faithful embedding of a category in the category of sets.  Definition~\ref{ic} generalizes both of these definitions, and was called a {\em quasicategory} in \citep{quasicats} and {\em weak Kan complexes} in \citep{weakkan} when ${\cal C}$ is a category.

\subsection{Example: Simplicial Objects over Integer-Valued Multisets} 

To help ground out this somewhat abstract discussion above on simplicial objects and sets, let us consider its application to two other examples.  Our first example comes from a  non-graphical representations of conditional independence, namely integer-valued multisets \citep{studeny2010probabilistic}, defined as an integer-valued multiset function $u: \mathbb{Z}^{{\cal P(\mathbb{Z})}} \rightarrow \mathbb{Z}$ from the power set of integers, ${\cal P(\mathbb{Z})}$ to integers $\mathbb{Z}$. An imset is defined over partialy ordered set (poset), defined as a distributive lattice of disjoint (or non-disjoint) subsets of variables. The bottom element is denoted $\emptyset$, and top element represents the complete set of variables $N$. A full discussion of the probabilistic representations induced by imsets is given \citep{studeny2010probabilistic}. We will only focus on the aspects of imsets that relate to its conditional independence structure, and its topological structure as defined by the poset.  A {\em combinatorial} imset is defined as: 

\[ u = \sum_{A \subset N} c_A \delta_A \]

where $c_A$ is an integer, $\delta_A$ is the characteristic function for subset $A$, and $A$ potentially ranges over all subsets of $N$. An {\em elementary} imset is defined over $(a,b \CI A)$, where $a,b$ are singletons, and $A \subset N \setminus \{a, b\}$. A {\em structural} imset is defined as one where the coefficients can be rational numbers. For a general DAG model $G = (V, E)$, an imset in standard form \citep{studeny2010probabilistic} is defined as 

\[ u_G = \delta_V - \delta_\emptyset + \sum_{i \in V} (\delta_{\mbox{{\bf  Pa}}_i} - \delta_{i \cup \mbox{{\bf Pa}}_i}) \] 

\begin{figure}[h] 
 \caption{An illustration of an integer-valued multiset (imset) consisting of a lattice of subsets over three elements for representing conditional independences in DAG models. All three DAG models are represented by the same imset. We can view an imset as a simplicial object (shown as the dark blue filled triangle) whose simplices map to integers.   \label{imset}} \vskip 0.1in
\centering
\begin{minipage}{0.5\textwidth}
\includegraphics[scale=0.3]{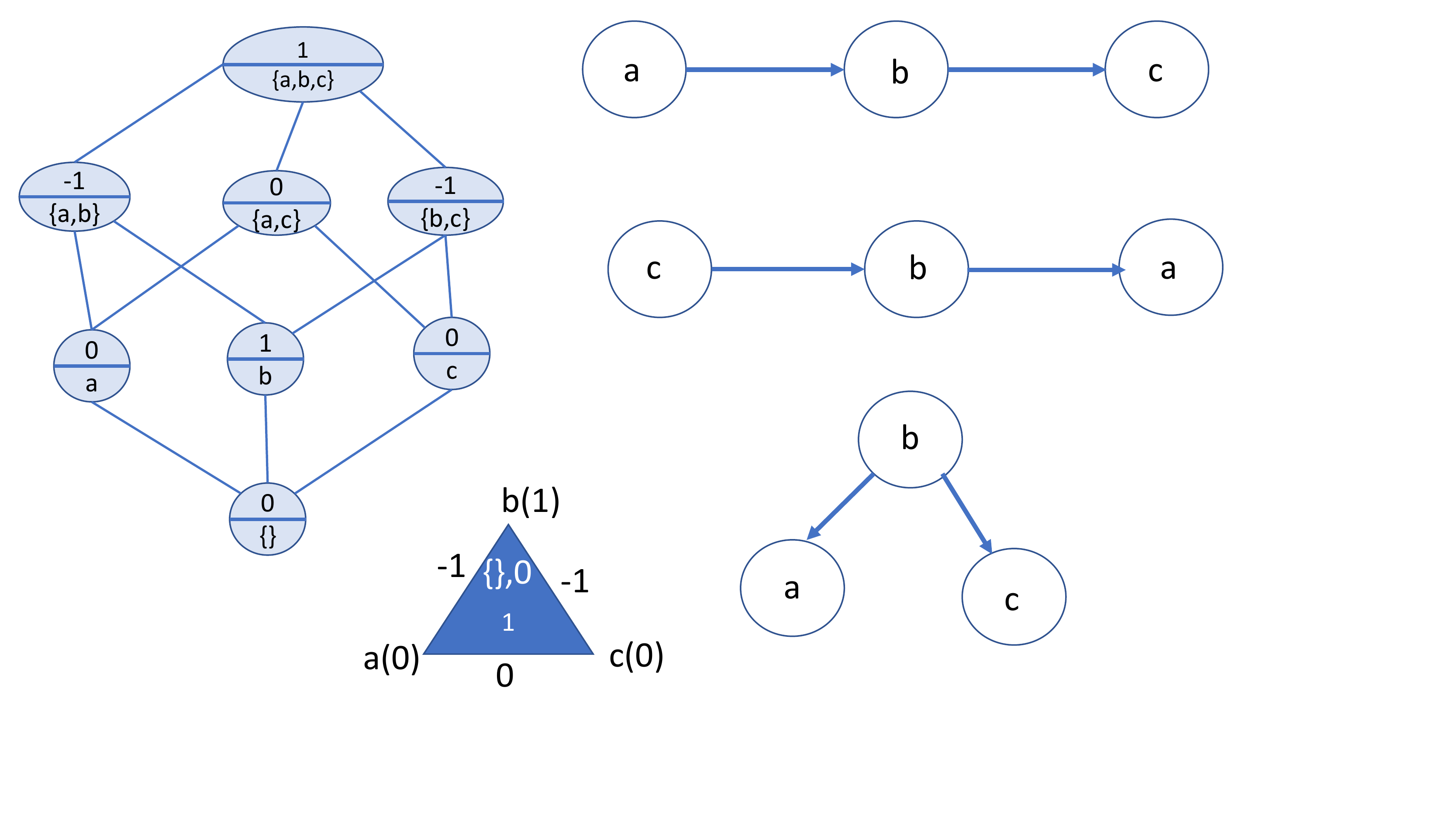}
\end{minipage} 
 \end{figure}

Figure~\ref{imset} shows an example imset for DAG models over three variables, defined by an integer valued function over the lattice of subsets. Each of the three DAG models shown defines exactly the same imset function.  \citet{studeny2010probabilistic} gives a detailed analysis of imsets as a non-graphical representation of conditional independence. 

\begin{figure}[h] 
 \caption{Causal discovery through the space of all models over $3$ variables shown with their associated imset representations \citep{imset-simplex}. Each candidate DAG defines a {\em causal horn}, a simplicial subobject of the complete simplex on $\Delta[2]$, and the process of causal structure discovery can be viewed in terms of the abstract horn filling problem defined above for higher-order categories. Each DAG is shown in terms of its ``essential graph", a reduced representation defined in \cite{anderson-annals}, where an undirected edge is used to denote the reversibility of a directed edge.  \label{imset-simp-obj}}
 \vskip 0.1in
\centering
\begin{minipage}{0.7\textwidth}
\includegraphics[scale=0.3]{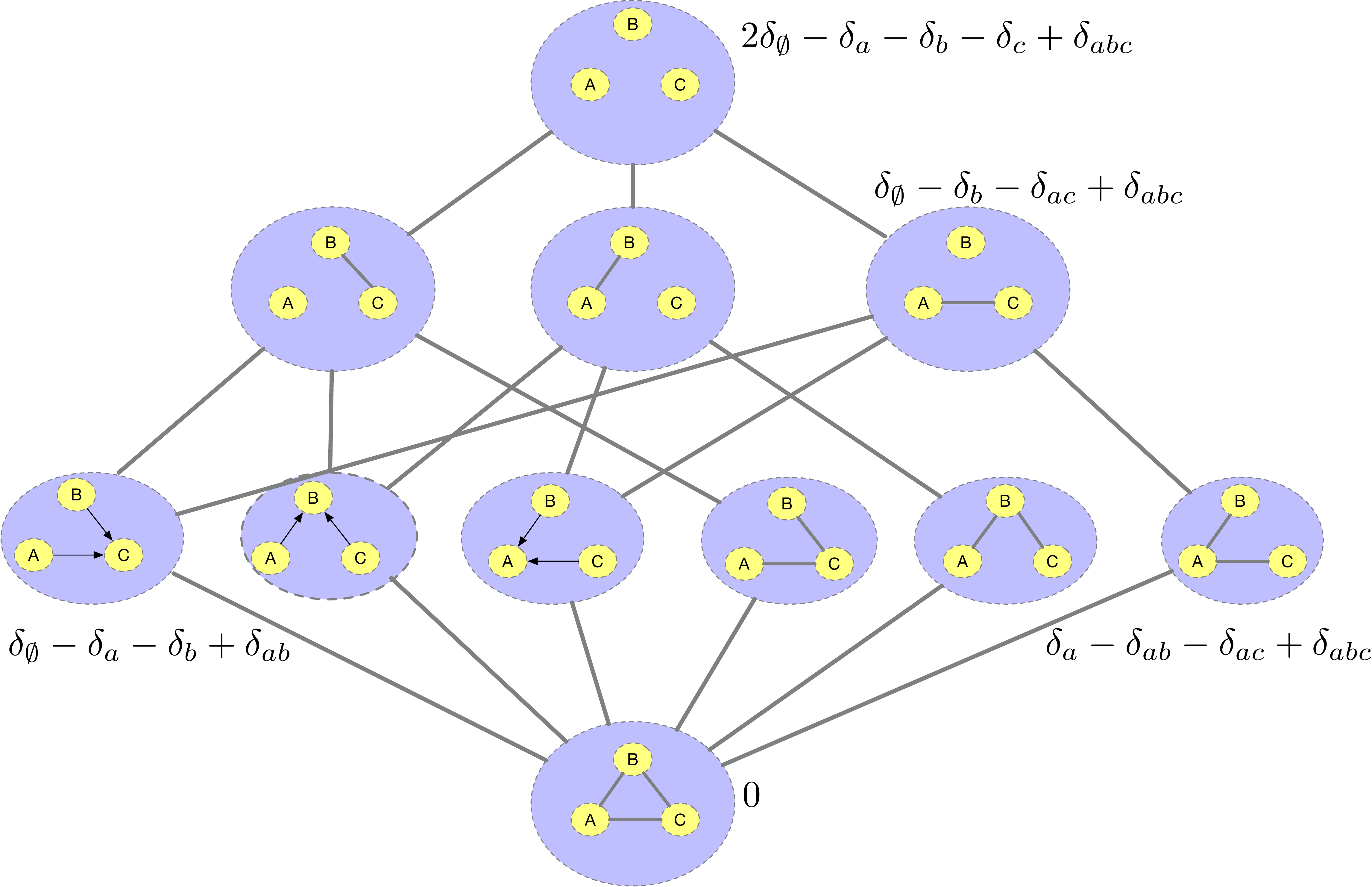}
\end{minipage} 
 \end{figure}

 Figure~\ref{imset-simp-obj} shows the space of possible imset representations over $3$ variables, along with the accompanying causal DAG shown in its reduced ``essential graph" form \cite{anderson-annals}. Each vertex in this figure shows a potential DAG model, along with its imset representation. We can now see how each of these DAG models can be viewed in terms of a simplicial object using functors that maps from the category $\Delta$ into this lattice of possible DAGs over $3$ variables. Note that the model at the very top can be essentially viewed as degenerate simplicial object, where all its arrows have been discarded.  Similiarly, the collider model $A \rightarrow C \leftarrow B$, whose integer-valued multiset representation is defined as $\delta_\emptyset - \delta_a - \delta_b + \delta_{ab}$, can be viewed as simplicial outer horn object $\Lambda^2_2$. Similarly, the serial DAG model $A \rightarrow B \rightarrow C$ can be viewed as the simplicial inner horn object $\Lambda^2_1$. 

\subsection{Example: Simplicial Objects over String Diagrams} 

\begin{figure}[h] 
\centering
\vskip 0.1in
\begin{minipage}{0.6\textwidth}
\includegraphics[scale=0.3]{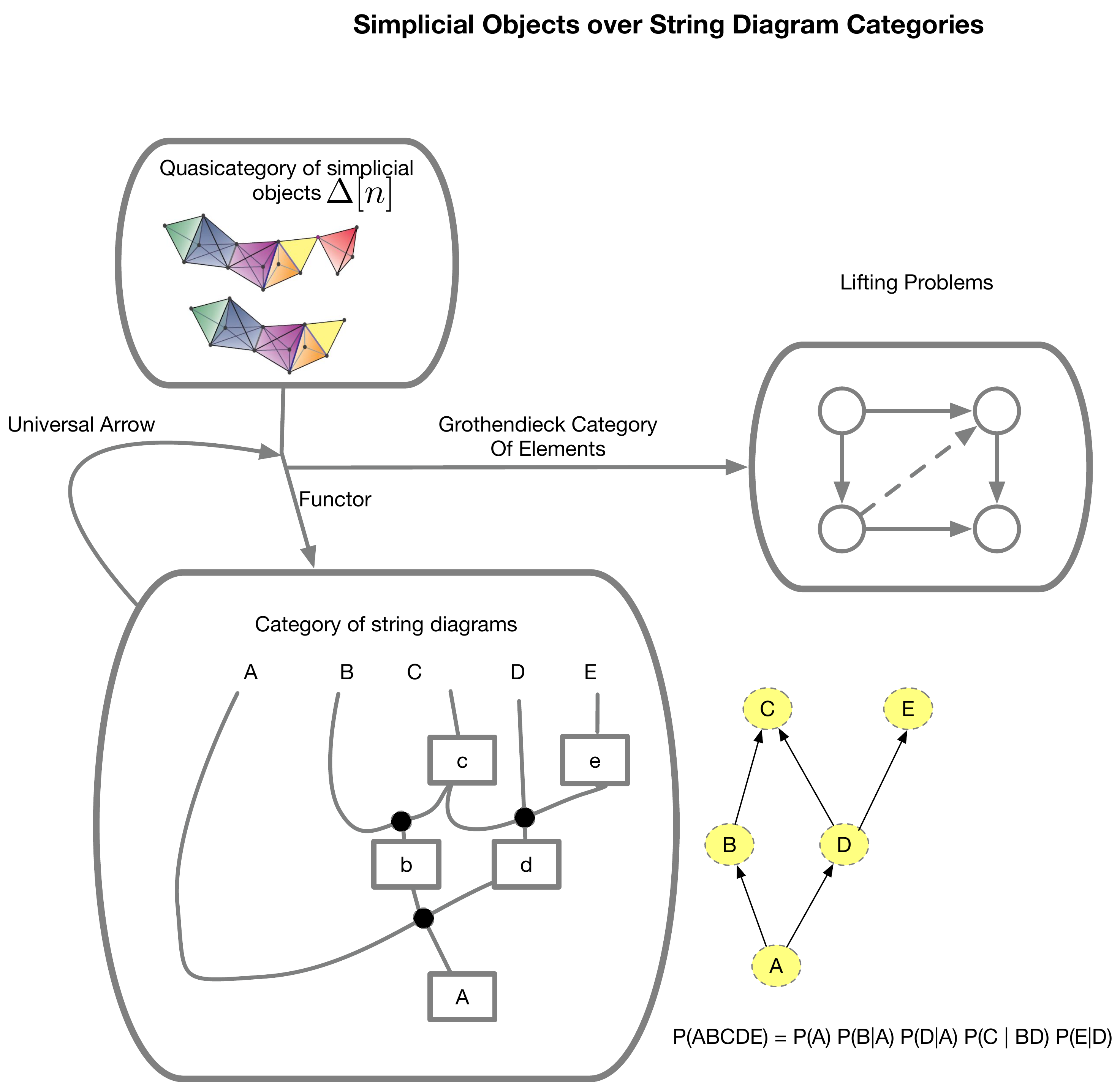}
\end{minipage}
\caption{Causal models over symmetric monoidal categories were explored by \citet{string-diagram-surgery}. \label{ucla-string} }
 \end{figure} 

We now illustrate the above formalism of simplicial objects by illustrating how it applies to the special case where causal models are defined over symmetric monoidal categories \cite{fong:ms,string-diagram-surgery}. For a detailed overview of symmetric monoidal categories, we recommend the book-length treatment by \citet{fong2018seven}. We will restrict ourselves to showing how the UCLA framework provides a way to generalize the past work on string diagrams for causal inference. Symmetric monoidal categories (SMCs) are useful in modeling processes where objects can be combined together to give rise to new objects, or where objects disappear. For example, \citet{Coecke_2016} propose a mathematical framework for resources based on SMCs. It is important to point out that monoidal categories can be defined as a special type of Grothendieck fibration \cite{richter2020categories}, thereby formally showing that the structure of a monoidal category is a special case of our UCLA hierarchy. We will discuss Grothendieck fibrations in more detail in the next section.  We leave aside the details of this construction, but refer the reader to \cite{richter2020categories} for the details.

We focus on the work of \citet{string-diagram-surgery}, and use the example shown in Figure~\ref{ucla-string}, where we have only shown the top two levels of the UCLA hierarchy. Our goal in this section is to illustrate how we can define simplicial objects over the SMC category CDU category {\bf Syn}$_G$ constructed by \citet{string-diagram-surgery} to mimic the process of working with an actual Bayesian network DAG $G$. An example of such an encoding is shown in Figure~\ref{ucla-string}. For the purposes of our illustration, it is not important to discuss the intricacies involved in this model, for which we refer the reader to the original paper. Our goal is to show that by encapsulating their SMC category in the UCLA framework, we can extend their approach as described below. In particular, we can solve an associated lifting problem that is defined by the functor mapping the simplicial category $\Delta$ to their SMC category. They use the category of stochastic matrices to capture the process of working with the joint distribution as shown in the figure. Instead, we show that one can use some other category, such as the category of {\bf Sets}, or {\bf Top} (the category of topological spaces), or indeed, the category {\bf Meas} of measurable spaces. 

Recall that Bayesian networks \cite{pearl:bnets-book} define a joint probability distribution 

 \[ P(X_1, \ldots, X_n) = \prod_{i=1}^n P(X_i | \mbox{Pa}(X_i)], \]  
 
 where $\mbox{Pa}(X_i) \subset \{X_1, \ldots, X_n \} \setminus {X_i}$ represents a subset of variables (not including the variable itself). \citet{string-diagram-surgery} show Bayesian network models can be constructed using symmetric monoidal categories, where the tensor product operation is used to combine multiple variables into a ``tensored" variable that then probabilistically maps into an output variable. In particular, the monoidal category {\bf Stoch} has as objects finite sets, and morphisms $f: A \rightarrow B$ are $|B| \times |A|$ dimensional stochastic matrices. Composition of stochastic matrices corresponds to matrix multiplication. The monoidal product $\otimes$ in {\bf Stoch} is the cartesian product of objects, and the Kronecker product of matrices $f \otimes g$. \citet{string-diagram-surgery} define three additional operations, the copy map, the discarding map, and the uniform state. 

 \begin{definition}
 A CDU category (for copy, discard, and uniform) is a SMC category ({\bf C}, $\otimes$, $I$), where each object $A$ has a copy map $C_A: A \rightarrow A \otimes A$, and discarding map $D_A: A \rightarrow I$, and a uniform state map $U_A: I \rightarrow A$, satisfying a set of equations detailed in \citet{string-diagram-surgery}. CDU functors are symmetric monoidal functors between CDU categories, preserving the CDU maps. 
 \end{definition}

 The key theorem we are interested in is the following from the original paper \cite{string-diagram-surgery}: 

 \begin{theorem}
There is an isomorphism (1-1 correspondence) between Bayesian networks based on a DAG $G$ and CDU functors $F:$ {\bf Syn}$_G \rightarrow$ {\bf Stoch}. 
 \end{theorem}

 The significance of this theorem for the UCLA architecture is that it shows how the SMC category of CDU objects can be defined as Layer 2 of the UCLA hierarchy, whereas the category {\bf Stoch} can be viewed as instantiating the Layer 3 of the UCLA hierarchy. 
 
 \subsection{Nerve of a Category}

 An important concept that will play a key role in Layer 4 of the UCLA hierarchy is that of the {\em nerve} of a category \cite{kerodon,richter2020categories}. The nerve of a category ${\cal C}$ enables embedding ${\cal C}$ into the category of simplicial objects, which is a fully faithful embedding. 

\begin{definition} 
\label{fully-faithful} 
Let ${\cal F}: {\cal C} \rightarrow {\cal D}$ be a functor from category ${\cal C}$ to category ${\cal D}$. If for all arrows $f$ the mapping $f \rightarrow F f$
\begin{itemize}
    \item injective, then the functor ${\cal F}$ is defined to be {\bf faithful}. 
    \item surjective, then the functor ${\cal F}$ is defined to be {\bf full}.  
    \item bijective, then the functor ${\cal F}$ is defined to be {\bf fully faithful}. 
\end{itemize}
\end{definition}

\begin{definition}
The {\bf nerve} of a category ${\cal C}$ is the set of composable morphisms of length $n$, for $n \geq 1$.  Let $N_n({\cal C})$ denote the set of sequences of composable morphisms of length $n$.  

\[ \{ C_o \xrightarrow[]{f_1} C_1 \xrightarrow[]{f_2} \ldots \xrightarrow[]{f_n} C_n \ | \ C_i \ \mbox{is an object in} \ {\cal C}, f_i \ \mbox{is a morphism in} \ {\cal C} \} \] 
\end{definition}

The set of $n$-tuples of composable arrows in {\cal C}, denoted by $N_n({\cal C})$,  can be viewed as a functor from the simplicial object $[n]$ to ${\cal C}$.  Note that any nondecreasing map $\alpha: [m] \rightarrow [n]$ determines a map of sets $N_m({\cal C}) \rightarrow N_n({\cal C})$.  The nerve of a category {\cal C} is the simplicial set $N_\bullet: \Delta \rightarrow N_n({\cal C})$, which maps the ordinal number object $[n]$ to the set $N_n({\cal C})$.

The importance of the nerve of a category comes from a key result \cite{kerodon}, showing it defines a full and faithful embedding of a category: 

\begin{theorem}
\cite[\href{https://kerodon.net/tag/002Y}{Tag 002Y}]{kerodon}: The {\bf nerve functor} $N_\bullet: {\bf Cat} \rightarrow {\bf Set}$ is fully faithful. More specifically, there is a bijection $\theta$ defined as: 

\[ \theta: {\bf Cat}({\cal C}, {\cal C'}) \rightarrow {\bf Set}_\Delta (N_\bullet({\cal C}), N_\bullet({\cal C'}) \] 
\end{theorem}

Using this concept of a nerve of a category, we can now state a theorem that shows it is possible to easily embed the CDU symmetric monoidal category defined above that represents Bayesian Networks and their associated ``string diagram surgery" operations for causal inference as a simplicial set. 

\begin{theorem}
  Define the {\bf nerve} of the CDU symmetric monoidal  category  ({\bf C}, $\otimes$, $I$), where each object $A$ has a copy map $C_A: A \rightarrow A \otimes A$, and discarding map $D_A: A \rightarrow I$, and a uniform state map $U_A: I \rightarrow A$ as the set of composable morphisms of length $n$, for $n \geq 1$.  Let $N_n({\cal C})$ denote the set of sequences of composable morphisms of length $n$.  

\[ \{ C_o \xrightarrow[]{f_1} C_1 \xrightarrow[]{f_2} \ldots \xrightarrow[]{f_n} C_n \ | \ C_i \ \mbox{is an object in} \ {\cal C}, f_i \ \mbox{is a morphism in} \ {\cal C} \} \]   

The associated {\bf nerve functor} $N_\bullet: {\bf Cat} \rightarrow {\bf Set}$ from the CDU category is fully faithful. More specifically, there is a bijection $\theta$ defined as: 

\[ \theta: {\bf Cat}({\cal C}, {\cal C'}) \rightarrow {\bf Set}_\Delta (N_\bullet({\cal C}), N_\bullet({\cal C'}) \] 
\end{theorem}

This theorem is just a special case of the above theorem attesting to the full and faithful embedding of any category using its nerve, which then makes it a simplicial set. We can then use the theoretical machinery at the top layer of the UCLA architecture to manipulate causal interventions in this category using face and degeneracy operators as defined above. 

Note that the functor $G$ from a simplicial object $X$ to a category ${\cal C}$ can be lossy. For example, we can define the objects of ${\cal C}$ to be the elements of $X_0$, and the morphisms of ${\cal C}$ as the elements $f \in X_1$, where $f: a \rightarrow b$, and $d_0 f = a$, and $d_1 f = b$, and $s_0 a, a \in X$ as defining the identity morphisms ${\bf 1}_a$. Composition in this case can be defined as the free algebra defined over elements of $X_1$, subject to the constraints given by elements of $X_2$. For example, if $x \in X_2$, we can impose the requirement that $d_1 x = d_0 x \circ d_2 x$. Such a definition of the left adjoint would be quite lossy because it only preserves the structure of the simplicial object $X$ up to the $2$-simplices. The right adjoint from a category to its associated simplicial object, in contrast, constructs a full and faithful embedding of a category into a simplicial set.  In particular, the  nerve of a category is such a right adjoint.

As this has been a rather long section, it is time to summarize and review what we have learned. We defined the top layer of the UCLA architecture as a category over ordinal numbers $\Delta$, and showed that the accompanying morphisms can be used to implement causal interventions using degeneracy and face operators. We illustrated using two examples -- the integer-valued multisets investigated by \citet{studeny2010probabilistic}, as well as the work on string diagram surgery by \citet{string-diagram-surgery} -- to help make concrete how the abstract machinery of simplicial objects and sets can be used to implement causal ``surgery" operators. We also discussed how categories can be faithfully embedded into simplicial sets using the nerve of the category, the set of sequences of lenth $n$ of composable morphisms. Next, we want to proceed downwards to explore the interaction between the causal layer and the layer of instances (or sets).

\section{Layers 2 and 3 of UCLA: The Category of Elements in Causal Inference}

\begin{figure}[h] 
\centering
\vskip 0.1in
\begin{minipage}{0.4\textwidth}
\includegraphics[scale=0.3]{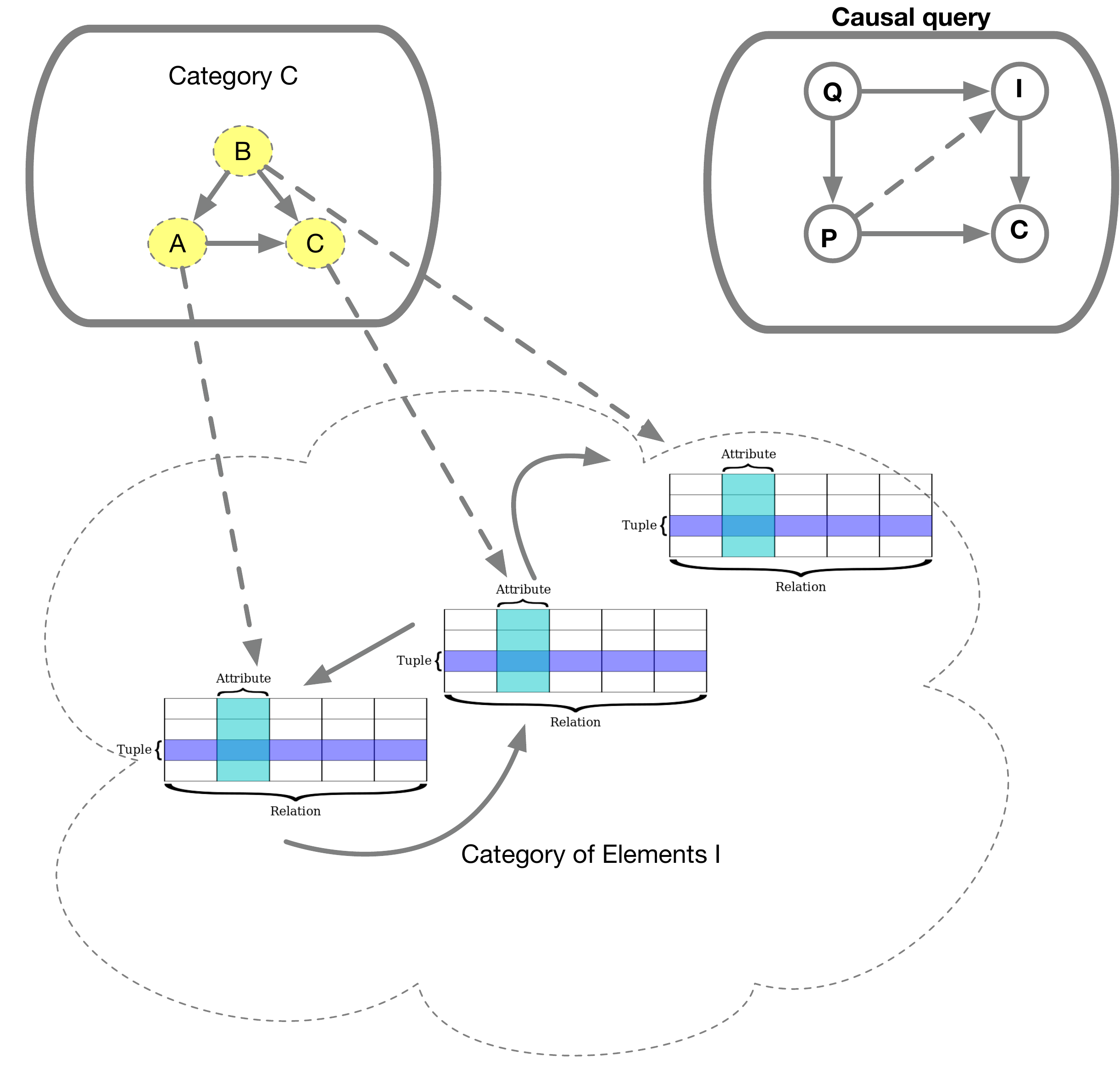}
\end{minipage}
\caption{Causal models can be viewed as defining a relational database over instances, where the database scheme defines a category. Causal queries can then be viewed analogously to database queries, and be formulated as solutions to lifting problems \cite{SPIVAK_2013}. \label{category-of-elements} }
 \end{figure}

Next, we turn to describe the second (from top) and third layers of the UCLA architecture, which pertain to the category of causal models (for example, a graph or a symmetric monoidal category), and the database of instances that support causal inferences. Figure~\ref{category-of-elements} illustrates the mapping from a causal category to a dataset of instances, modeled as a category {\bf Set}. Drawing on the close correspondences between between categories and relational database schemes (see \cite{SPIVAK_2013} for details), we can view causal queries over data as analogous to database queries, which can then be formulated by corresponding lifting problems. In this figure, the lifting problem with respect to the category $I$ of elements, which combines the causal model with the instances that it maps into. That is, each object in the  model, e.g. a variable indicating a patient, maps into actual patients, and a variable indicating outcomes from Covid-19 exposure, maps into actual outcomes for that individual. The causal arrow from the patient variable into the exposure variable then maps into actual arrows for each patient. Causal queries of exposure to Covid-19 then become similar to database queries. In the next section, we will generalize this perspective, showing that we can map into a topological category and answer more abstract questions relating to the geometry of a dataset, or map into a category of measurable spaces to answer probabilistic queries. The structure of the lifting problem remains the same, what changes are the specifics of the underlying categories. 

A central and unifying principle in UC and the UCLA architecture is that every pair of categorical layers is synchronized by a functor, along with a universal arrow.  We explore the universal arrow property more deeply in this section, showing how it provides the conceptual basis behind  the Yoneda Lemma, and Grothendieck's category of elements. In the case of causal inference, universal arrows enable mimicking the effects of causal operations from one layer of the UCLA hierarchy down to the next layer. In particular, at the simplicial object layer, we can model a causal intervention in terms of face and degeneracy operators (defined below in more detail). These in turn correspond to ``graph surgery" \cite{pearl:causalitybook} operations on causal DAGs, or in terms of ``copy", ``delete" operators in ``string diagram surgery" of causal models defined on symmetric monoidal categories \cite{string-diagram-surgery}. These ``surgery" operations at the next level may translate down to operations on probability distributions, measurable spaces, topological spaces, or chain complexes. This process follows a standard construction used widely in mathematics, for example group representations associate with any group $G$, a left {\bf k}-module $M$ representation that enables modeling abstract group operations by operations on the associated modular representation.  These concrete representations must satisfy the universal arrow property for them to be faithful.  

A special case of the universal arrow property is that of universal element, which as we will see below plays an important role in the UCLA architecture in defining a suitably augmented category of elements, based on a construction introduced by Grothendieck. 

\begin{definition}
If $D$ is a category and $H: D \rightarrow {\bf Set}$ is a set-valued functor, a {\bf universal element} associated with the functor $H$ is a pair $\langle r, e \rangle$ consisting of an object $r \in D$ and an element $e \in H r$ such that for every pair $\langle d, x \rangle$ with $x \in H d$, there is a unique arrow $f: r \rightarrow d$ of $D$ such that $(H f) e = x$. 
\end{definition}

\begin{example}
Let $E$ be an equivalence relation on a set $S$, and consider the quotient set $S/E$ of equivalence classes, where $p: S \rightarrow S/E$ sends each element $s \in S$ into its corresponding equivalence class. The set of equivalence classes $S/E$ has the property that any function $f: S \rightarrow X$ that respects the equivalence relation can be written as $f s = f s'$ whenever $s \sim_E s'$, that is, $f = f' \circ p$, where the unique function $f': S/E \rightarrow X$. Thus, $\langle S/E, p \rangle$ is a universal element for the functor $H$. 
\end{example}

 \subsection{The Category of Elements}

We turn next to define the category of elements, based on a construction by Grothendieck, and illustrate how it can serve as the basis for inference at each layer of the UCLA architecture. In particular, \citet{SPIVAK_2013} shows how the category of elements can be used to define SQL queries in a relational database. 

\begin{definition}
Given a set-valued functor $\delta: {\cal C} \rightarrow {\bf Set}$ from some category ${\cal C}$, the induced {\bf category of elements} associated with $\delta$ is a pair $(\int \delta, \pi_\delta)$, where $\int \delta \in$ {\bf Cat} is a category in the category of all categories {\bf Cat}, and $\pi_\delta: \int \delta \rightarrow {\cal C}$ is a functor that ``projects" the category of elements into the corresponding original category ${\cal C}$. The objects and arrows of $\int \delta$ are defined as follows: 
\begin{itemize}
    \item $\mbox{Ob}(\int \delta) = \{ (s, x) | x \in \mbox{Ob}({\cal c}), x \in \delta s \} $. 

    \item {\bf Hom}$_{\int \delta}((s, x), (s', x')) = \{f: s \rightarrow s' | \delta f (x) = x' \}$
\end{itemize}
\end{definition}

\begin{figure}[h] 
 \caption{A toy causal DAG model of climate change to illustrate the category of elements construction.  \label{climate-change}}
 \vskip 0.1in
\centering
\begin{minipage}{0.7\textwidth}
\includegraphics[scale=0.3]{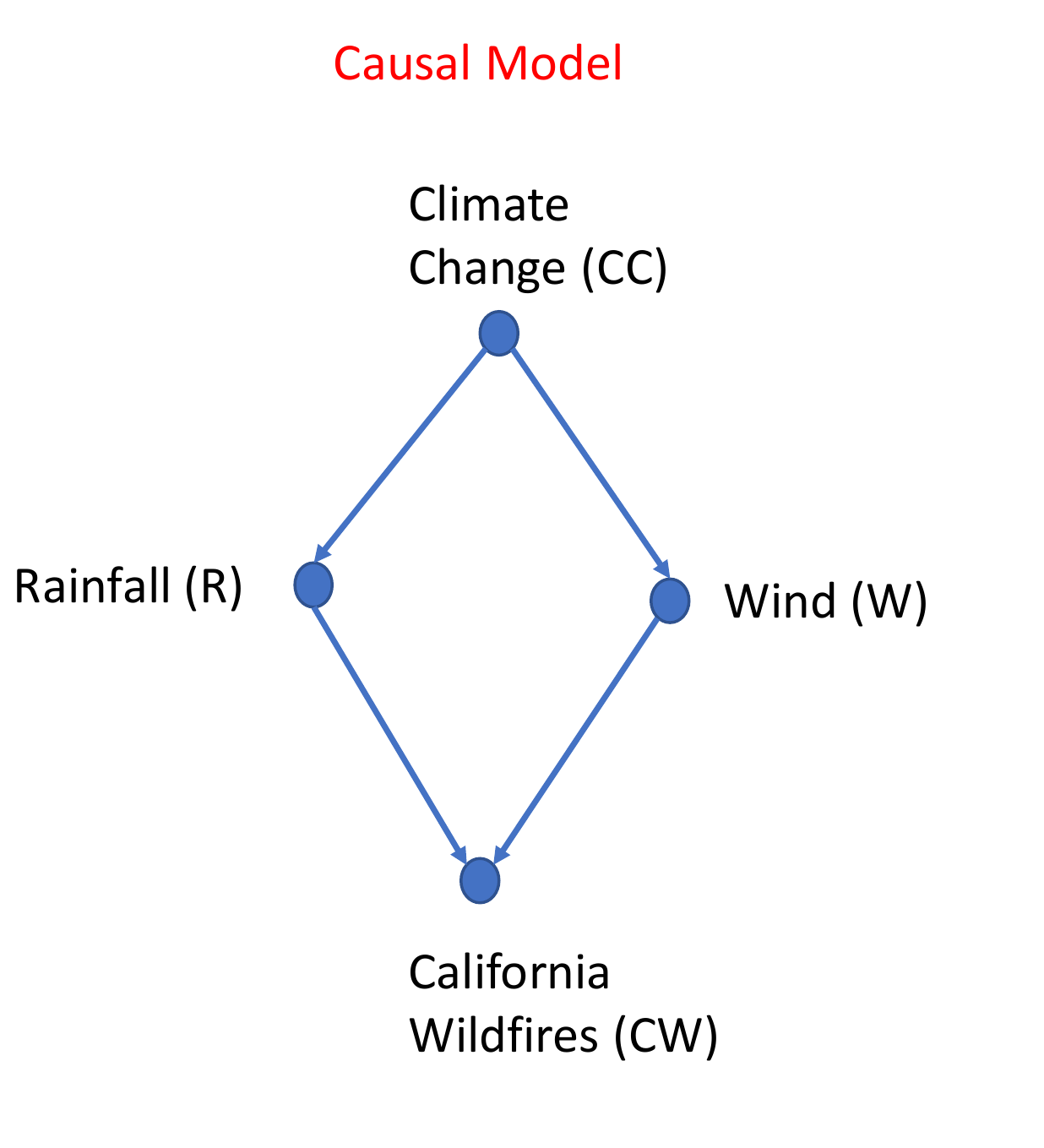}
\includegraphics[scale=0.3]{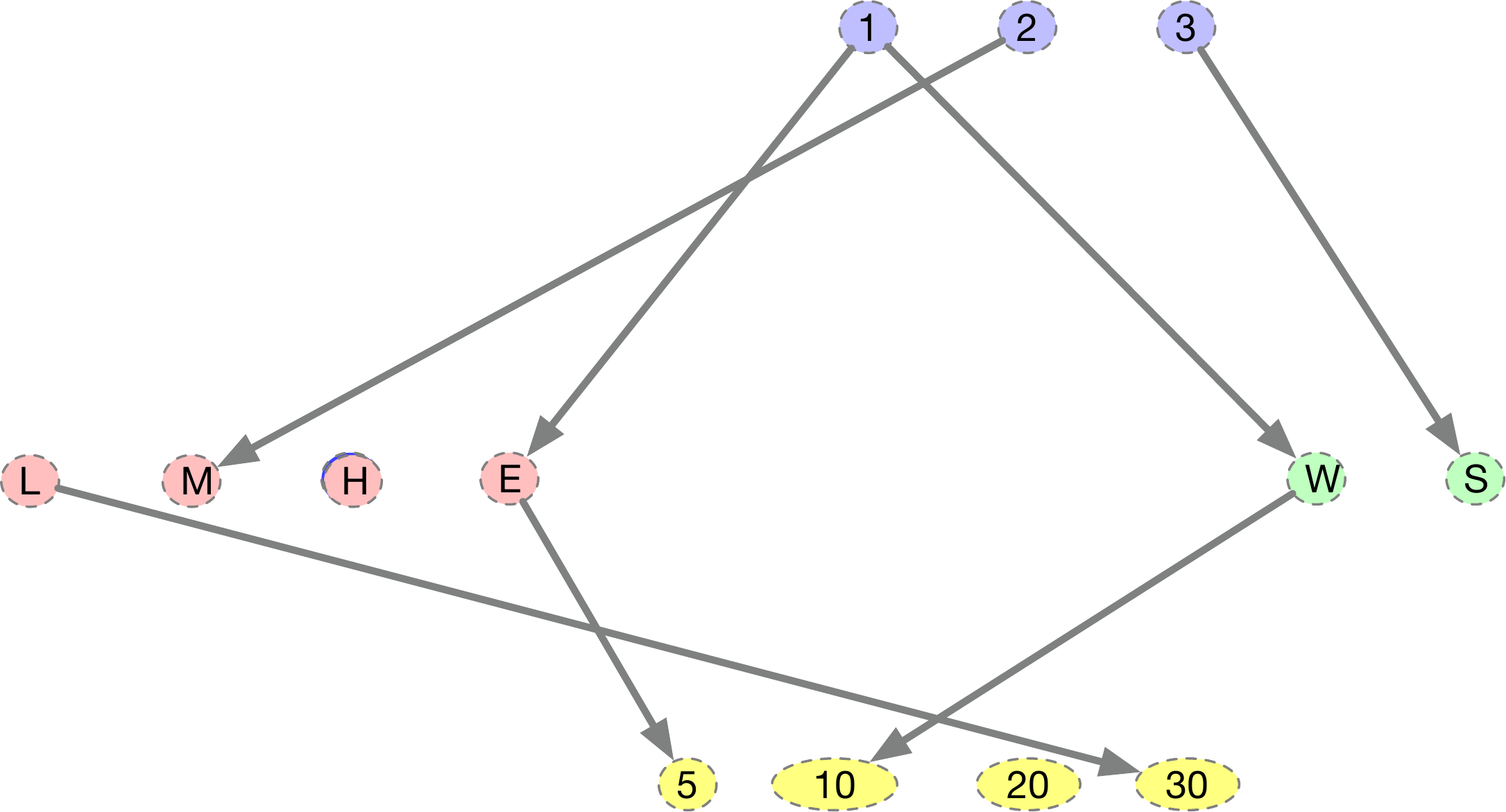}
\end{minipage} 
 \end{figure}

\begin{example}
To illustrate the category of elements construction, let us consider the toy climate change causal model shown in Figure~\ref{climate-change}. Let the category {\cal C} be defined by this causal DAG model, where the objects Ob({\cal C}) are defined by the four vertices, and the arrows {\bf Hom}$_{\cal C}$ are defined by the four edges in the model. The set-valued functor $\delta: {\cal C} \rightarrow {\bf Set}$ maps each object (vertex) in {\cal C} to a set of instances, thereby turning the causal DAG model into an associated set of tables. For example, {\bf Climate Change} is defined as a table of values, which could be modeled as a multinomial variable taking on a set of discrete values, and for each of its values, the arrow from {\bf Climate Change} to {\bf Rainfall} maps each specific value of {\bf Climate Change} to a value of {\bf Rainfall}, thereby indicating a causal effect of climate change on the amount of rainfall in California. Im the figure, {\bf Climate Change} is mapped to three discrete levels (marked $1$, $2$ and $3$). Rainfall amounts are discretized as well into low (marked "L"), medium (marked "M"), high (marked "H"), or extreme (marked "E"). Wind speeds are binned into two levels (marked "W" for weak, and "S" for strong). Finally, the percentage of California wildfires is binned between $5$ to $30$. Not all arrows that exist in the Grothendieck category of elements are shown, for clarity. 

 \end{example}

 Many properties of Grothendieck's construction can be exploited (some of these are discussed in the context of relational database queries in \cite{SPIVAK_2013}), but for our application to causal inference, we are primarily interested in the associated class of lifting problems that define queries in a causal model. In particular, it can be shown that Grothendieck's category of elements formally defines a type of fibration called a {\em Grothendieck opfibration}. More specifically, if we consider any object $c \in {\cal C}$ in a category ${\cal C}$, and look at its fiber $\pi_\delta^{-1}(c)$, which forms a subcategory of the category of elements $I$, it can be shown that the projection of every morphism in the fiber into the category ${\cal C}$ is to the identity element ${\bf 1}_c$. \citet{richter2020categories} explains how monoidal categories themselves can be defined using simplicial objects using a Grothendieck opfibration. We refer the reader to these sources for additional details.  One consequence of this property is that approaches to UC that build symmetric monoidal category representations of Bayesian networks \cite{fong:ms,string-diagram-surgery} can be formally shown to define special cases of the UCLA hierarchy. 
 
\begin{definition}
 If $S$ is a collection of morphisms in category  ${\cal C}$, a morphism $f: A \rightarrow B$ has the {\bf left lifting property with respect to S} if it has the left lifting property with respect to every morphism in $S$. Analogously, we say a morphism $p: X \rightarrow Y$ has the {\bf right lifting property with respect to S} if it has the right lifting property with respect to every morphism in $S$.
 \end{definition}

 \subsection{Lifting Problems in Causal Inference}

We now turn to sketch some examples of the application of lifting problems for causal inference. Many problems in causal inference on graphs involve some particular graph property.  To formulate it as a lifting problem, we will use the following generic template, following the initial application of lifting problems to database queries proposed by \citet{SPIVAK_2013}.

\begin{center}
 \begin{tikzcd}
  Q \arrow{d}{f} \arrow{r}{\mu}
    & \int \delta  \arrow[red]{d}{p} \\
  R \arrow[ur,dashed, "h"] \arrow[red]{r}[blue]{\nu}
&{\cal C} \end{tikzcd}
 \end{center} 

 Here, $Q$ is a generic query that we want answered, which could range from a database query, as in the original setting studied by \citet{SPIVAK_2013}, but more interestingly, it could be a particular graph property relating to causal inference (as illustrated by the following two examples), but as we will show later, it could also be related to the combinatorial category of simplicial objects used to model causal intervention, and finally, it could also be related to questions relating to the evaluation of causal models using a measure-theoretic or probability space. By suitably modifying the base category, the lifting problem formulation can be used to encode a diverse variety of problems in causal inference. $R$ represents a fragment of the complete causal model ${\cal C}$, and $\delta$ is the category of elements defined above. Finally, $h$ gives all solutions to the lifting problem. Some examples will help clarify this concept.

\begin{example}
 Consider the category of directed graphs defined by the category ${\cal G}$, where Ob(${\cal G}$) = \{V, E\}, and the morphisms of ${\cal G}$ are given as {\bf Hom}$_{\cal G}$ = \{s, t\}, where $s: E \rightarrow V$ and $t: E \rightarrow V$ define the source and terminal nodes of each vertex. Then, the category of all directed graphs is precisely defined by the category of all functors $\delta: {\cal G} \rightarrow {\bf Set}$. Any particular graph is defined by the functor $X: {\cal G} \rightarrow {\bf Set}$, where the function $X(s): X(E) \rightarrow X(V)$ assigns to every edge its source vertex. For causal inference, we may want to check some property of a graph, such as the property that every vertex in $X$ is the source of some edge. The following lifting problem ensures that every vertex has a source edge in the graph. The category of elements $\int \delta$ shown below refers to a construction introduced by Grothendieck, which will be defined in more detail later. 
 
 \begin{center}
 \begin{tikzcd}
  V (\bullet) \arrow{d}{f} \arrow{r}{\mu}
    & \int \delta  \arrow[red]{d}{p} \\
  \{ E (\bullet) \xrightarrow[]{s} V (\bullet) \} \arrow[ur,dashed, "h"] \arrow[red]{r}[blue]{\nu}
&{\cal G} \end{tikzcd}
 \end{center} 

 \end{example}

\begin{example} 

As another example of the application of lifting problems to causal inference, let us consider the problem of determining whether two causal DAGs, $G_1$ and $G_2$ are Markov equivalent \cite{anderson-annals}. A key requirement here is that the immoralitiies of $G_1$ and $G_2$ must be the same, that is, if $G_1$ has a collider $A \rightarrow B \leftarrow C$, where there is no edge between $A$ and $C$, then $G_2$ must also have the same collider, and none others. We can formulate the problem of finding colliders as the following lifting problem. Note that the three vertices $A$, $B$ and $C$ are bound to an actual graph instance through the category of elements $\int \delta$ (as was illustrated above), using the top right morphism $\mu$. The bottom left morphism $f$ binds these three vertices to some collider. The bottom right morphism $\nu$ requires this collider to exist in the causal graph ${\cal G}$ with the same bindings as found by $\mu$. The dashed morphsm $h$ finds all solutions to this lifting problem, that is, all colliders involving the vertices $A$, $B$ and $C$. 
 
 \begin{center}
 \begin{tikzcd}
  \{A (\bullet), B (\bullet), C (\bullet) \} \arrow{d}{f} \arrow{r}{\mu}
    & \int \delta  \arrow[red]{d}{p} \\
  \{ A (\bullet) \rightarrow B (\bullet) \leftarrow  C (\bullet)\} \arrow[ur,dashed, "h"] \arrow[red]{r}[blue]{\nu}
&{\cal G} \end{tikzcd}
 \end{center} 

 \end{example}

If the category of elements is defined by a functor mapping a database schema into a table of instances, then the associated lifting problem corresponds to familiar problems like SQL queries in relational databases \cite{SPIVAK_2013}. In our application, we can use the same machinery to formulate causal inference queries by choosing the categories appropriately.  To complete the discussion, we now make the connection between universal arrows and the core notion of universal representations via the Yoneda Lemma.

\subsection{Modeling Causal Interventions as Kan Extension}

 It is well known in category theory that ultimately every concept, from products and co-products, limits and co-limits, and ultimately even the Yoneda Lemma (see below), can be derived as special cases of the Kan extension \citep{maclane:71}. Kan extensions intuitively are a way to approximate a functor ${\cal F}$ so that its domain can be extended from a category ${\cal C}$ to another category  ${\cal D}$.  Because it may be impossible to make commutativity work in general, Kan extensions rely on natural transformations to make the extension be the best possible approximation to ${\cal F}$ along ${\cal K}$.  We want to briefly show Kan extensions can be combined with the category of elements defined above to construct causal ``migration functors" that map from one causal model into another. These migration functors were originally defined in the context of database migration \cite{SPIVAK_2013}, and here we are adapting that approach to causal inference. By suitably modifying the category of elements from a set-valued functor $\delta: {\cal C} \rightarrow {\bf Set}$, to some other category, such as the category of topological spaces, namely $\delta: {\cal C} \rightarrow {\bf Top}$, we can extend the causal migration functors into solving more abstract causal inference questions. We explore the use of such constructions in the next section on Layer 4 of the UCLA hierarchy. Here, for simplicity, we restrict our focus to Kan extensions for migration functors over the category of elements defined over instances of a causal model. 

\begin{definition}
A {\bf left Kan extension} of a functor $F: {\cal C} \rightarrow {\cal E}$ along another functor $K: {\cal C} \rightarrow {\cal D}$, is a functor $\mbox{Lan}_K F: {\cal D} \rightarrow {\cal E}$ with a natural transformation $\eta: F \rightarrow \mbox{Lan}_F \circ K$ such that for any other such pair $(G: {\cal D} \rightarrow {\cal E}, \gamma: F \rightarrow G K)$, $\gamma$ factors uniquely through $\eta$. In other words, there is a unique natural transformation $\alpha: \mbox{Lan}_F \implies G$. \\
%
\begin{center}
\begin{tikzcd}[row sep=2cm, column sep=2cm]
\mathcal{C}  \ar[dr, "K"', ""{name=K}]
            \ar[rr, "F", ""{name=F, below, near start, bend right}]&&
\mathcal{E}\\
& \mathcal{D}  \ar[ur, bend left, "\text{Lan}_KF", ""{name=Lan, below}]
                \ar[ur, bend right, "G"', ""{name=G}]
                
%
\arrow[Rightarrow, "\exists!", from=Lan, to=G]
\arrow[Rightarrow, from=F, to=K, "\eta"]
\end{tikzcd}
\end{center}
\end{definition}

A {\bf right Kan extension} can be defined similarly.  To understand the significance of Kan extensions for causal inference, we note that under a causal intervention, when a causal category $S$ gets modified to $T$, evaluating the modified causal model over a database of instances can be viewed as an example of Kan extension. 

First, we need to review the basic concept of adjoint functors, which will be helpful in seeing how to use Kan extensions to model causal interventions. 

\begin{definition}
A pair of {\bf adjoint functors} is defined as $F: {\cal C}\rightarrow {\cal D}$ and $G: {\cal D} \rightarrow {\cal C}$, where $F$ is considered the right adjoint, and $G$ is considered the left adjoint, 

\[
        \begin{tikzcd}
            {\cal D} \arrow[r, shift left=1ex, "G"{name=G}] & \C\arrow[l, shift left=.5ex, "F"{name=F}]
            \arrow[phantom, from=F, to=G, , "\scriptscriptstyle\boldsymbol{\top}"].
        \end{tikzcd}
    \]

must satisfy the property that for each pair of objects $C$ of ${\cal C}$ and $D$ of ${\cal D}$, there is a bijection of sets 

\[ \phi_{C, D}: \mbox{{\bf Hom}}_{\cal C}(C, G(D)) \simeq \mbox{{\bf Hom}}_{\cal D}(F(C), D)  \]

\end{definition} 

Notice the similarity of this definition to the one earlier where the universal arrow property induced a bijection of {\bf Hom} sets that then led to universal elements, Grothendieck category of elements, and the Yoneda Lemma. 

\begin{figure}[h] 
 \caption{Kan extensions are useful in modeling the effects of a causal intervention.  \label{kan-causal-intervention}}
 \vskip 0.1in
\centering
\begin{minipage}{0.5\textwidth}
\includegraphics[scale=0.3]{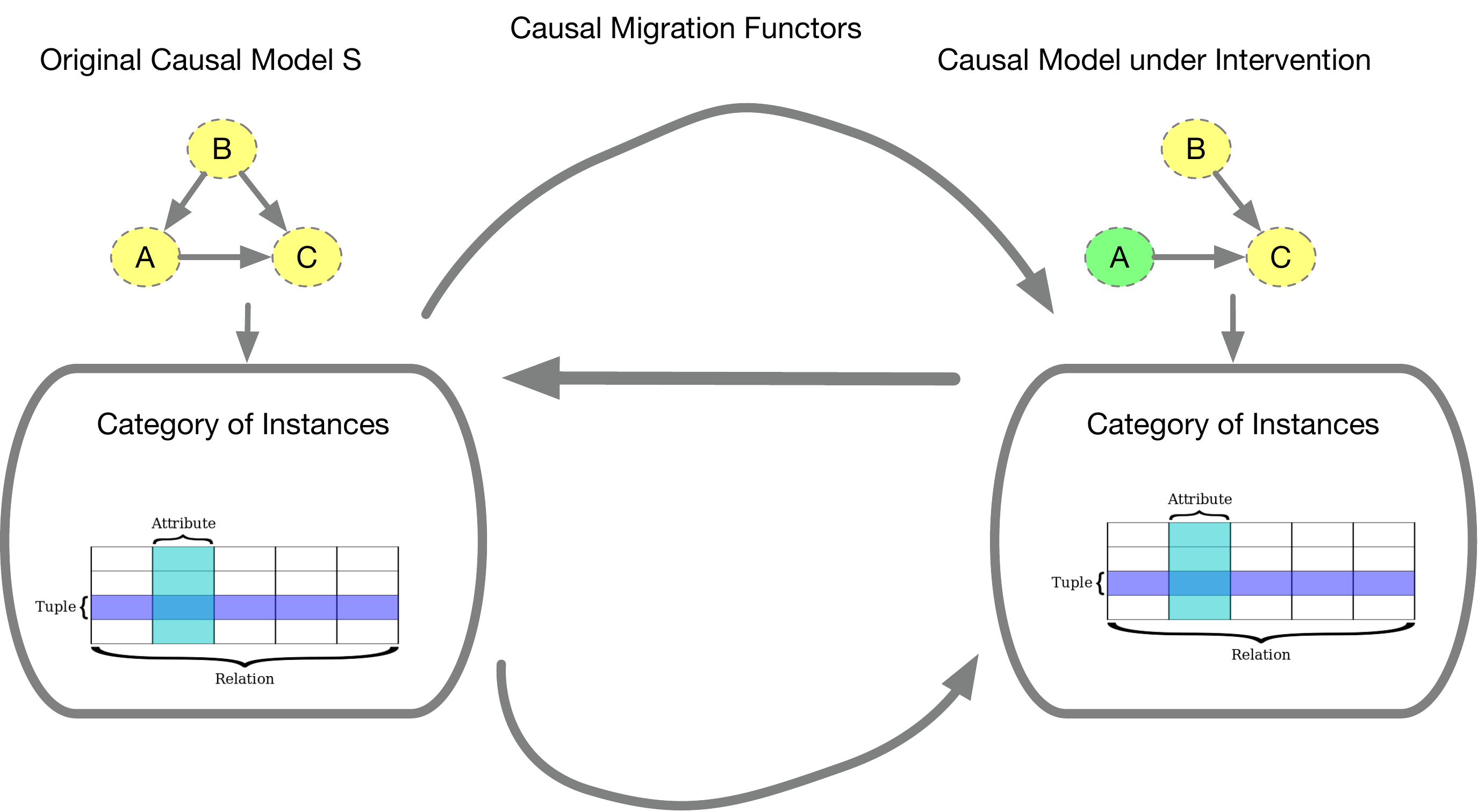}
\end{minipage} 
 \end{figure}

 Let $\delta: S \rightarrow {\bf Set}$ denote the original causal model defined by the category $S$ with respect to some dataset. Let $\epsilon: T \rightarrow {\bf Set}$ denote the effect of a causal intervention abstractly defined as some change in the category $S$ to $T$, such as deletion of an edge, as illustrated in Figure~\ref{kan-causal-intervention}. Intuitively, we can consider three cases: the {\em pullback} $\Delta_F$ along $F$, which maps the effect of a causal intervention back to the original model, the {\em left pushforward} $\Sigma_F$ and the {\em right pushforward} $\prod_F$, which can be seen as adjoints to the pullback $\Delta_F$.

Following \cite{SPIVAK_2013}, we can define three {\em causal migration functors} that evaluate the impact of a causal intervention with respect to a dataset of instances.

\begin{enumerate} 
\item The functor $\Delta_F: \epsilon \rightarrow \delta$ sends the functor $\epsilon: T \rightarrow {\bf Set}$ to the composed functor $\delta \circ F: S \rightarrow {\bf Set}$. 

\item The functor $\Sigma_F: \delta \rightarrow \epsilon$ is the left Kan extension along $F$, and can be seen as the left adjoint to $\Delta_F$. 

The functor $\prod_F: \delta \rightarrow \epsilon$ is the right Kan extension along $F$, and can be seen as the right adjoint to $\Delta_F$. 
\end{enumerate}

To understand how to implement these functors, we use the following proposition that is stated in \cite{SPIVAK_2013} in the context of database queries, which we are restating in the setting of causal inference. 

\begin{theorem}
Let $F: S \rightarrow T$ be a functor. Let $\delta: S \rightarrow {\bf Set}$ and $\epsilon: T \rightarrow {\bf Set}$ be two set-valued functors, which can be viewed as two instances of a causal model defined by the category $S$ and $T$. If we view $T$ as the causal category that results from a causal intervention on $S$ (e.g., deletion of an edge), then there is a commutative diagram linking the category of elements between $S$ and $T$. 

\begin{center}
 \begin{tikzcd}
  \int \delta \arrow{d}{\pi_\delta} \arrow{r}{}
    & \int \epsilon \arrow[red]{d}{\pi_\epsilon} \\
  S \arrow[red]{r}[blue]{F}
&T \end{tikzcd}
 \end{center} 

\end{theorem}

{\bf Proof:} To check that the above diagram is a pullback, that is, $\int \delta \simeq S \times_T \int \delta$, or in words, the fiber product, we can check the existence of the pullback component wise by comparing the set of objects and the set of morphisms in $\int \delta$ with the respective sets in $S \times_T \int \epsilon$. $\bullet$

For simplicity, we defined the migration functors above with respect to an actual dataset of instances. More generally, we can compose the set-valued functor $\delta: S \rightarrow {\bf Set}$ with a functor ${\cal T}: {\bf Set} \rightarrow {\bf Top}$ to the category of topological spaces to derive a Kan extension formulation of the definition of a causal intervention. We discuss this issue in the next section on causal homotopy.

\subsection{Yoneda Lemma}

The Yoneda Lemma plays a crucial role in UC because it defines the concept of a representation in category theory. We first show that associated with universal arrows is the corresponding induced isomorphisms between {\bf Hom} sets of morphisms in categories. This universal property then leads to the Yoneda Lemma. 

\begin{theorem}
Given any functor $S: D \rightarrow C$, the universal arrow $\langle r, u: c \rightarrow Sr \rangle$ implies a bijection exists between the {\bf Hom} sets 
\[ \mbox{{\bf Hom}}_{D}(r, d) \simeq \mbox{{\bf Hom}}_{C}(c, Sd) \]
\end{theorem}

This is a well-known result whose proof can be found in \cite{maclane:71}. The crucial point here can be illustrated with the help of a commutative diagram, showing that the bijection is defined by a natural transformation $\phi$, natural in the object $d$ in category $D$. By investigating a special case of this natural transformation for how it transforms the identity morphism {\bf 1}$_r$ leads us to the Yoneda Lemma. 

\begin{center}
 \begin{tikzcd}
  D(r,r) \arrow{d}{D(r, f')} \arrow{r}{\phi_r}
    & C(c, Sr) \arrow[red]{d}{C(c, S f')} \\
  D(r,d)  \arrow[red]{r}[blue]{\phi_d}
&C(c, Sd)\end{tikzcd}
 \end{center} 

Note here that the identity morphism {\bf 1}$_r \in D(r,r)$ is mapped to $S f' \circ u$ through the top and right path, and to $\phi_d(f')$ through the left and bottom path. As these paths must be equal in a commutative diagram, we get the ensuring property that a bijection between the {\bf Hom} sets holds precisely when $\langle r, u: c \rightarrow Sr \rangle$ is a universal arrow from $c$ to $S$. Note that for the case when the categories $C$ and $D$ are small, meaning their {\bf Hom} collection of arrows forms a set, the induced functor {\bf Hom}$_C(c, S - )$ to {\bf Set} is isomorphic to the functor {\bf Hom}$_D(r, -)$. This type of isomorphism defines a (universal) representation. 

\begin{lemma}
{\bf Yoneda Lemma)}: If $H: D \rightarrow \mbox{{\bf Set}}$ is a set-valued functor, and $r$ is an object in $D$, there is a bijection that sends each natural transformation $\alpha: \mbox{{\bf Hom}}_D(r, -) \rightarrow K$ to $\alpha_r \mbox{{\bf 1}}_r$, the image of the identity morphism {\bf 1}$_r: r \rightarrow r$. 
\[ y: \mbox{{\bf Nat}}(\mbox{{\bf Hom}}_D(r, -), K) \simeq K r \]
\end{lemma}

The proof of the Yoneda Lemma follows directly from the below commutative diagram, a special case of the above diagram for universal arrows. 

\begin{center}
 \begin{tikzcd}
  D(r,r) \arrow{d}{D(r, f')} \arrow{r}{\phi_r}
    & K r \arrow[red]{d}{C(c, S f')} \\
  D(r,d)  \arrow[red]{r}[blue]{\phi_d}
&K d \end{tikzcd}
 \end{center} 

 \subsection{The Universality of Diagrams in Causal Inference}

 Diagrams play a key role in defining UC and the UCLA architecture, as has already become clear from the discussion above. We briefly want to emphasize the central role played by universal constructions involving limits and colimits of diagrams, which are viewed as functors from an indexing category of diagrams to a category. To make this somewhat abstract definition concrete, let us look at some simpler examples of universal properties, including co-products and quotients (which in set theory correspond to disjoint unions). Coproducts refer to the universal property of abstracting a group of elements into a larger one.

\begin{center}
\begin{tikzcd}
    & Z\arrow[r, "p"] \arrow[d, "q"]
      & X \arrow[d, "f"] \arrow[ddr, bend left, "h"]\\
& Y \arrow[r, "g"] \arrow[drr, bend right, "i"] &X \sqcup Y \arrow[dr, "r"]  \\ 
& & & R 
\end{tikzcd}
\end{center} 

In the commutative diagram above, the coproduct object $X \sqcup Y$ uniquely factorizes any mapping $h: X \rightarrow R$ and any mapping $i: Y \rightarrow R$, so that $h = r \circ f$, and furthermore $i = r \circ g$. Co-products are themselves special cases of the more general notion of co-limits. 

\subsubsection{Pullback Mappings} 

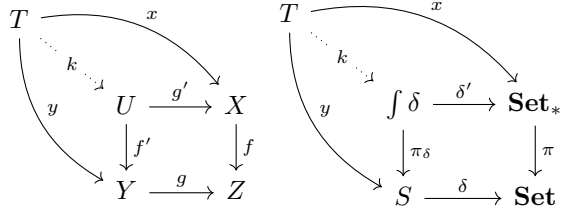
\begin{figure}[h]
\centering
\begin{tikzcd}
  T
  \arrow[drr, bend left, "x"]
  \arrow[ddr, bend right, "y"]
  \arrow[dr, dotted, "k" description] & & \\
    & 
    U\arrow[r, "g'"] \arrow[d, "f'"]
      & X \arrow[d, "f"] \\
& Y \arrow[r, "g"] &Z
\end{tikzcd}
\begin{tikzcd}
  T
  \arrow[drr, bend left, "x"]
  \arrow[ddr, bend right, "y"]
  \arrow[dr, dotted, "k" description] & & \\
    & 
    \int \delta \arrow[r, "\delta'"] \arrow[d, "\pi_\delta"]
      & {\bf Set}_* \arrow[d, "\pi"] \\
& S \arrow[r, "\delta"] & {\bf Set}
\end{tikzcd}
\caption{Left: Universal Property of pullback mappings. Right: The Grothendieck category of elements $\int \delta$ of any set-valued functor $\delta: S \rightarrow {\bf Set}$ can be described as a pullback in the diagram of categories.} 
\label{univpr}
\end{figure}

Figure~\ref{univpr}  illustrates the fundamental property of a {\em pullback}, which along with {\em pushforward}, is one of the core ideas in category theory. The pullback square with the objects $U,X, Y$ and $Z$ implies that the composite mappings $g \circ f'$ must equal $g' \circ f$. In this example, the morphisms $f$ and $g$ represent a {\em pullback} pair, as they share a common co-domain $Z$. The pair of morphisms $f', g'$ emanating from $U$ define a {\em cone}, because the pullback square ``commutes" appropriately. Thus, the pullback of the pair of morphisms $f, g$ with the common co-domain $Z$ is the pair of morphisms $f', g'$ with common domain $U$. Furthermore, to satisfy the universal property, given another pair of morphisms $x, y$ with common domain $T$, there must exist another morphism $k: T \rightarrow U$ that ``factorizes" $x, y$ appropriately, so that the composite morphisms $f' \ k = y$ and $g' \ k = x$. Here, $T$ and $U$ are referred to as {\em cones}, where $U$ is the limit of the set of all cones ``above" $Z$. If we reverse arrow directions appropriately, we get the corresponding notion of pushforward. So, in this example, the pair of morphisms $f', g'$ that share a common domain represent a pushforward pair. 
As Figure~\ref{univpr}, for any set-valued functor $\delta: S \rightarrow {\bf Sets}$, the Grothendieck category of elements $\int \delta$ can be shown to be a pullback in the diagram of categories. Here, ${\bf Set}_*$ is the category of pointed sets, and $\pi$ is a projection that sends a pointed set $(X, x \in X)$ to its underlying set $X$.

To state the next two results, we need to introduce formally the concept of limit and colimits \cite{maclane:71}. 

\begin{definition}
Given a functor $F: {\cal J} \rightarrow {\cal C}$ from an indexing diagram category ${\cal J}$ to a category ${\cal C}$, an element $A$ from the set of natural transformations $N(A,F)$ is called a {\bf cone}. A {\bf limit} of the diagram $F: {\cal J} \rightarrow {\cal C}$ is a cone $\eta$ from an object lim $F$ to the diagram satisfying the universal property that for any other cone $\gamma$ from an object $B$ to the diagram, there is a unique morphism $h: B \rightarrow \mbox{lim} F$ so that $\gamma \bullet = \eta \bullet h$ for all objects $\bullet$ in ${\cal J}$. Dually, the {\bf co-limit} of the diagram $F: {\cal J} \rightarrow {\cal C}$ is a cone $\epsilon$ satisfying the universal property that for any other cone $\gamma$ from the diagram to the object $B$, there is a unique mapping $h: \mbox{colim} F \rightarrow B$ so that $\gamma \bullet = h \epsilon \bullet$ for all objects $\bullet$ in ${\cal J}$. 
\end{definition}

A fundamental consequence of the category of elements is that every object in the functor category of presheaves, namely contravariant functors from a category into the category of sets, is the colimit of a diagram of representable objects, via the Yoneda Lemma. This is a standard result whose proof can be found in \cite{maclane:sheaves}, and will be of importance below. 

\begin{theorem}
\label{presheaf-theorem}
In the functor category of presheaves {\bf Set}$^{{\cal C}^{op}}$, every object $P$ is the colimit of a diagram of representable objects, in a canonical way. 
\end{theorem}

In UC, any causal influence of an object $X$ upon any other object $Y$ can be represented as a natural transformation (a morphism) between two functor objections in the presheaf category $\hat{{\cal C}}$. The CRP is very akin to the idea of the reproducting property in kernel methods. Reproducing Kernel Hilbert Spaces (RKHS's)  transformed the study of machine learning, precisely because they are the unique subcategory in the category of all Hilbert spaces that have representers of evaluation defined by a kernel matrix $K(x,y)$ \cite{kernelbook}. The reproducing property in an RKHS is defined as $\langle K(x, -), K(-, y) \rangle = K(x,y)$. An analogous but far more general reproducing property holds in the UC framework, based on the Yoneda Lemma. 

\begin{theorem}
\label{crp}
{\bf Causal Reproducing Property:}  All causal influences between any two objects $X$ and $Y$ can be derived from its presheaf functor objects, namely 

\[ \mbox{{\bf Hom}}_{\cal C}(X,Y) \simeq \mbox{{\bf Nat}}(\mbox{{\bf Hom}}_{\cal C}(-, X),{\mbox{\bf Hom}}_{\cal C}(-, Y)) \]
\end{theorem}

{\bf Proof:} The proof of this theorem is a direct consequence of the Yoneda Lemma, which states that for every presheaf functor object $F$ in  $\hat{{\cal C}}$ of a category ${\cal C}$, {\bf Nat}({\bf Hom}$_{\cal C}(-, X), F) \simeq F X$. That is, elements of the set $F X$ are in $1-1$ bijections with natural transformations from the presheaf {\bf Hom}$_{\cal C}(-, X)$ to $F$. For the special case where the functor object $F = $ {\bf Hom}$_{\cal C}(-, Y)$, we get the result immediately that  {\bf Hom}$_{\cal C}(X,Y) \simeq$ {\bf Nat}({\bf Hom}$_{\cal C}(-, X)$,{\bf Hom}$_{\cal C}(-, Y))$. $\bullet$

The significance of the Causal Reproducing Property is that presheaves act as ``representers" of causal information, precisely analogous to how kernel matrices act as representers in an RKHS.

\section{Layer 4 of UCLA: Evaluating Causal Effects using Homotopy Colimits}  

Finally, we turn to discuss the role of the causal homotopy layer. Before delving into the details of this category, we pause to review  what is meant by ``causal effect". Previous work in causal inference has adopted a wide variety of definitions of causality. For example, in the work on potential outcomes \cite{rubin-book}, a causal effect is defined through the average treatment effect (ATE) estimator, where the mean of the outcomes $E(Y_1)$ under treatment is compared with the mean of the outcomes of the outcomes under control $E(Y_0)$. In the work on DAG-based causal inference \citep{pearl:causalitybook}, causal effects are defined as a change in the distribution of values of the treated variable $P(Y | X = 1)$ compared to the untreated variable $P(Y | X = 0)$. \citet{janzing} propose an information-theoretic approach to quantifying causal influence, where the causal effect is defined in terms of the KL-divergence between the observational $P(Y | X = x)$ and interventional distributions  $P(Y | do(X = x))$. Work on reproducing kernel Hilbert space embeddings of causal models, such as mean counterfactual embeddings \cite{cme}, propose evaluating the ATE estimator in the induced RKHS, so that in the original data space, the estimator is defined with respect to a reproducing kernel. 

Since our approach to causality is based on category theory, we prefer to derive an estimator that is entirely based on the structure of categories, namely objects and morphisms. In particular, we defined the nerve of the category above as a full and faithful embedding of a category as a simplicial object. Associated with the nerve is an important topological invariant of a category called its {\em classifying space} \cite{richter2020categories}, which provides a way of defining algebraic invariants with a causal model. 

\subsection{Intuition Underlying Our Definition of Causal Effects}

Before proceeding to give rigorous definitions, it is useful to build up some intuition. Evaluating causal effects in terms of probabilistic models, or statistical models, or information-theoretic measures is so widespread that it may be difficult to imagine other ways of thinking about this problem. Let us consider what it means to evaluate the average treatment effect in a clinical trial, e.g., patients being tested for a Covid-19 vaccine, or indeed the counterfactual evaluation of the efficacy of a Covid-19 vaccine on patients in a hospital suffering from the effects of Covid-19. Consider the causal pathways that occur for patients who did take the vaccine, vs. the pathways for patients who did not take the vaccine. For the latter group, those who ended up in the hospital will end up suffering potential long-term effects from the Covid-19 infection, incur other medical complications such as shortness of breath, and end up requiring long-term treatment. Patients who were vaccinated and did not require hospitalizations would not in general require some medical intervention. In the population at large, of course, there are bound to be patients who despite being vaccinated end up in the hospital and end up requiring possibly long-term care, and similarly, patients who were never vaccinated still do not end up being infected by Covid-19 at a level serious enough to require hospitalization. 

However, given the widespread published statistics showing patients admitted to hospitals for serious Covid-19 infections are preponderantly those who were not vaccinated, it is clear there is sufficient statistical regularity to discriminate between the two groups using an ATE estimator (either for testing a drug, or counterfactually, to determine whether they indeed would have required hospitalizations had they been vaccinated). 

Using our formulation in terms of the category of elements, we can discriminate between two sets of causal pathways through the space, one requiring hospitalizations and long-term care, and the other who do not. The statistical regularity that discriminates between vaccinated and non-vaccinated patients also manifests itself in a change in the underlying density of trajectories between the two cases. The topological model we propose below can uncover the difference in these densities, and provides a non-statistical approach to evaluating causal effects. 

For example, the method UMAP (uniform manifold approximation and projection) \cite{mcinnes2018uniform} is a powerful dimensionality reduction and data visualization method that is derived using the framework of category theory. UMAP can embed data in a lower-dimensional space in a way that separates objects from one class from objects in another class. UMAP relies on simplicial objects in terms of its design (actually, it uses an extended notion of ``fuzzy" simplicial sets, where each simplicial set is mapped to a locale $[0,1]$, the unit interval). Thus, by using such a category-theoretic or topologically aware method, one can discriminate between the classes of objects that were intervened on from those who were not.

\subsection{Singular Homology} 

Before we describe our approach to defining causal effect more rigorously,   we need to first define how to construct the topological realization of a category. We will then explore the properties of the classifying space of a category, and show how it can be useful in quantifying causal effects. We will also clarify how it relates to determining equivalences among causal models, namely homotopical invariance, and also how it sheds light on causal identification. 

First, we need to define more concretely the topological $n$-simplex that provides a concrete way to attach a topology to a simplicial object. Our definitions below build on those given in \citep{kerodon}. For each integer $n$, define the topological space $|\Delta_n|$ realized by the object $\Delta_n$ as 

\[ |\Delta_n| = \{t_0, t_1, \ldots, t_n  \in \mathbb{R}^{n+1}: t_0 + t_1 + \ldots + t_n = 1 \} \] 

This is the familiar $n$-dimensional simplex over $n$ variables. For any causal model, its classifying space $|{\cal N}_\bullet({\cal C})|$ defines a topological space. We can now define the {\em singular} $n$-simplex as a continuous mapping $\sigma: |\Delta_N| \rightarrow |{\cal N}_\bullet({\cal C})|$. Every singular $n$-simplex $\sigma$ induces a collection of $n-1$-dimensional simplices called {\em faces}, denoted as 

\[ d_i \sigma(t_0, \ldots, t_{n-1}) = (t_0, t_1, \ldots, t_{i-1}, 0, t_i, \ldots, t_{n-1}) \] 

Note that as discussed above, a causal intervention on a variable in a DAG can be modeled as applying one of these degeneracy operators $d_i$. The above definition shows that every such intervention has an effect on the topology associated with the causal model. Define the set of all morphisms $\mbox{Sing}_n(X) = {\bf Hom}_{\bf Top}(\Delta_n, |{\cal N}_\bullet({\cal C})|)$ as the set of singular $n$-simplices of $|{\cal N}_\bullet({\cal C})|$. 

\begin{definition}
For any topological space defined by a causal model $|{\cal N}_\bullet({\cal C})|$,  the {\bf singular homology groups} $H_*(|{\cal N}_\bullet({\cal C})|; {\bf Z})$ are defined as the homology groups of a chain complex 

\[ \ldots \xrightarrow[]{\partial} {\bf Z}(\mbox{Sing}_2(|{\cal N}_\bullet({\cal C})|)) \xrightarrow[]{\partial} {\bf Z}(\mbox{Sing}_1(|{\cal N}_\bullet({\cal C})|)) \xrightarrow[]{\partial} {\bf Z}(\mbox{Sing}_0(|{\cal N}_\bullet({\cal C})|)) \] 

where ${\bf Z}(\mbox{Sing}_n(|{\cal N}_\bullet({\cal C})|))$ denotes the free Abelian group generated by the set $\mbox{Sing}_n(|{\cal N}_\bullet({\cal C})|)$ and the differential $\partial$ is defined on the generators by the formula 

\[ \partial (\sigma) = \sum_{i=0}^n (-1)^i d_i \sigma \] 
\end{definition}

Intuitively, a chain complex builds a sequence of vector spaces that can be used to construct an algebraic invariant of a PSR from its classifying space  by choosing the left {\bf k} module ${\bf Z}$ to be a vector space. Each differential $\partial$ then becomes a linear transformation whose representation is constructed by modeling its effect on the basis elements in each ${\bf Z}(\mbox{Sing}_n(X))$. 

\begin{example}
Let us illustrate the singular homology groups defined by an integer-valued multiset \cite{studeny2010probabilistic} used to model conditional independence. Imsets over a DAG of three variables $N = \{a, b, c \} $ shown previously as Figure~\ref{imset} can be viewed as a finite discrete topological space. For this topological space $X$, the singular homology groups $H_*(X; {\bf Z})$ are defined as the homology groups of a chain complex 

\[  {\bf Z}(\mbox{Sing}_3(X)) \xrightarrow[]{\partial}  {\bf Z}(\mbox{Sing}_2(X)) \xrightarrow[]{\partial} {\bf Z}(\mbox{Sing}_2(X)) \xrightarrow[]{\partial} {\bf Z}(\mbox{Sing}_1(X)) \xrightarrow[]{\partial} {\bf Z}(\mbox{Sing}_0(X)) \] 

where ${\bf Z}(\mbox{Sing}_i(X))$ denotes the free Abelian group generated by the set $\mbox{Sing}_i(X)$ and the differential $\partial$ is defined on the generators by the formula 

\[ \partial (\sigma) = \sum_{i=0}^4 (-1)^i d_i \sigma \] 

The set $\mbox{Sing}_n(X)$ is the set of all morphisms ${\bf Hom}_{Top}(|\Delta_n|, X)$. For an imset over the three variables $N = \{a, b, c \}$, we can define the singular $n$-simplex $\sigma$ as: 

\[ \sigma: |\Delta^4| \rightarrow X \ \ \mbox{where} \ \ |\Delta^n | = \{t_0, t_1, t_2, t_3 \in [0,1]^4 : t_0 + t_1 + t_2 + t_3 = 1 \} \] 

The $n$-simplex $\sigma$ has a collection of faces denoted as $d_0 \sigma, d_1 \sigma, d_2 \sigma$ and $ d_3 \sigma$.  If we pick the $k$-left module ${\bf Z}$ as the vector space over real numbers $\mathbb{R}$, then the above chain complex represents a sequence of vector spaces that can be used to construct an algebraic invariant of a topological space defined by the integer-valued multiset.  Each differential $\partial$ then becomes a linear transformation whose representation is constructed by modeling its effect on the basis elements in each ${\bf Z}(\mbox{Sing}_n(X))$. An alternate approach to constructing a chain homology for an integer-valued multiset is to use M\"obius inversion to define the chain complex in terms of the nerve of a category (see our recent work on categoroids \citep{categoroids} for details). 
\end{example}

\subsection{Classifying Spaces and Homotopy Colimits}

Building on the intuition proposed above, we now introduce a formal way to define causal effects in our framework, which relies on the construction of a topological space associated with the nerve of a category. As we saw above, the nerve of a category is a full and faithful embedding of a category as a simplicial object. 

\begin{definition}
The {\bf classifying space} of a causal  model defined as a category ${\cal C}$ is the topological space associated with the nerve of the category $|N_\bullet {\cal C}|$. 
\end{definition}

We now want to bring in the set-valued functor mapping each causal category ${\cal C}$ to the actual experiment used, e.g., in a clinical trial \cite{rubin-book}, to evaluate average treatment effect or quantify the effect of a {\bf do} calculus intervention. We can then compute the topological space prior to intervention, and subsequent to intervention, and compare the two topological spaces in terms of their algebraic invariants (e.g., the chain complex, as described below). 

In general, we may want to evaluate a causal model not only with respect to the data used in a causal experiment, but also with respect to some underlying topological space or some measurable space. We can extend the above definition straightforwardly to these cases using an appropriate functor ${\cal T}: {\bf Set} \rightarrow {\bf Top} $, or alternatively ${\cal M}: {\bf Set} \rightarrow {\bf Meas}$. These augmented constructions can then be defined with respect to a more general notion called the {\em homotopy colimit} \cite{richter2020categories} of a causal model. 

\begin{definition}
The {\bf  homotopy colimit} $\mbox{hocolim}_{{\cal T} \circ \delta}$ of a causal  model  associated with a category ${\cal C}$, along with its associated category of elements associated with  a set-valued functor $\delta: {\cal C} \rightarrow {\bf Set}$, and a topological functor ${\cal T}: {\bf Set} \rightarrow {\bf Top}$ is isomorphic to topological space associated with the nerve of the category of elements, that is  $\mbox{hocolim}_{{\cal T} \circ \delta} \simeq|N_\bullet \left(\int \delta \right) |$. 
\end{definition}

To understand the classifying space $|N_\bullet {\cal C}|$ of a causal model defined as a category ${\cal C}$, let us go over some simple examples to gain some insight. 

\begin{example}
For any set $X$, which can be defined as a discrete category ${\cal C}_X$ with no non-trivial morphisms, the classifying space $|N_\bullet {\cal C}_X|$ is just the discrete topology over $X$ (where the open sets are all possible subsets of $X$). 
\end{example}

\begin{example} 
If we take a causal model defined as a partially ordered set $[n]$, with its usual order-preserving morphisms, then the nerve of $[n]$ is isomorphic to the representable functor $\delta(-, [n])$, as shown by the Yoneda Lemma, and in that case, the classifying space is just the topological space $\Delta_n$ defined above. 
\end{example}

\begin{example}
The classifying space $|N_\bullet {\cal C}_{CDU}|$ associated with CDU symmetric monoidal  category  encoding of a causal Bayesian DAG (see Figure~\ref{ucla-string}) is defined using the monoidal category ({\bf C}, $\otimes$, $I$), where each object $A$ has a copy map $C_A: A \rightarrow A \otimes A$, and discarding map $D_A: A \rightarrow I$, and a uniform state map $U_A: I \rightarrow A$, is defined as the topological realization of its nerve.   As before, the nerve $N_n({\cal C})$ of the CDU category is defined as the set of sequences of composable morphisms of length $n$.  

\[ \{ C_o \xrightarrow[]{f_1} C_1 \xrightarrow[]{f_2} \ldots \xrightarrow[]{f_n} C_n \ | \ C_i \ \mbox{is an object in} \ {\cal C}, f_i \ \mbox{is a morphism in} \ {\cal C} \} \]   

Note that the CDU category was associated with a CDU functor $F:$ {\bf Syn}$_G \rightarrow$ {\bf Stoch} to the category of stochastic matrices. We can now define the homotopy colimit $\mbox{hocolim}_{{\cal F}}$ of the CDU causal  model  associated with the CDU category ${\cal C}$, along with its associated category of elements associated with  a set-valued functor $\delta: {\cal C} \rightarrow {\bf Set}$, and a topological functor ${\cal F}: {\bf Set} \rightarrow {\bf Stoch}$ is isomorphic to topological space associated with the nerve of the category of elements over the composed functor, that is  $\mbox{hocolim}_{{\cal F} \circ \delta}$. 

\end{example}

\subsection{Defining Causal Effect}

Finally, we turn to defining causal effect using the notion of classifying space and homotopy colimits, as defined above. Space does not permit a complete discussion of this topic, but the basic idea is that once a causal model is defined as a topological space, there are a large number of ways of comparing two topological spaces from analyzing their chain complexes, or using a topological data analysis method such as UMAP \cite{mcinnes2018uniform}. 

In conventional approaches, such as potential outcomes, one defines a random variable $Y$ for outcomes (e.g., of a clinical trial), and then compares the difference of the outcomes for patients that were administered a drug vs. patients who were administered a placebo, which is formally written as $E(Y(1) - Y(0))$, where the expectation is over the population of all patients, and requires counterfactual estimation of quantities that are not in the data (e.g., patients who took the drug have no known outcomes for their outcomes under the placebo, and vice versa). In our UCLA framework, the causal effect between the two corresponding treatment and no-treatment cases amounts to a comparison of the classifying spaces. We formally define causal effect as follows. 

\begin{definition}
Let the classifying space under ``treatment" be defined as the topological space $|N_\bullet {\cal C}_1|$ associated with the nerve of category ${\cal C}_1$ under some intervention, which may result in a topological deformation of the model (e.g., deletion of an edge). Similarly, the classifying space under ``no treatment" be defined as the $|N_\bullet {\cal C}_0|$ under a no-treatment setting, with no intervention. A {\bf causally non-isomorphic effect} exists between categories ${\cal C}_1$ and ${\cal C}_0$, or ${\cal C}_1 \not \cong {\cal C}_0$ if and only if there is no invertible morphism $f: |N_\bullet {\cal C}_1| \rightarrow N_\bullet ({\cal C}_0|$ between the ``treatment" and ``no-treatment" topological spaces,  namely $f$ must be both {\em left invertible} and {\em right invertible}. 
\end{definition}

There is an equivalent notion of causal effect using the homotopy colimit definition proposed above, which defines the  nerve functor using the category of elements. This version is particularly useful in the context of evaluating a causal model over a dataset. 

\begin{definition}
Let the   homotopy colimit $\mbox{hocolim}_{1} = |N_\bullet(\int \delta_1)|$ be the topological space associated with a causal  category ${\cal C}_1$ under the ``treatment' condition be defined with respect to an associated category of elements defined by  a set-valued functor $\delta_1: {\cal C} \rightarrow {\bf Set}$ over a dataset of ``treated" variables, and corresponding ``no-treatment" $\mbox{hocolim}_{0} = |N_\bullet(\int \delta_0)|$  be the topological space of a causal  model  associated with a category ${\cal C}_0$ be defined over an associated category of elements defined by  a set-valued functor $\delta_0: {\cal C} \rightarrow {\bf Set}$ over a dataset of ``placebo" variables. A {\bf causally non-isomorphic effect} exists between categories ${\cal C}_1$ and ${\cal C}_0$, or ${\cal C}_1 \not \cong {\cal C}_0$ if and only if there is no invertible morphism $f: |N_\bullet (\int \delta_1)| \rightarrow |N_\bullet (\delta_0)|$ between the ``treatment" and ``no-treatment" homotopy colimit topological spaces,  namely $f$ must be both {\em left invertible} and {\em right invertible}.
\end{definition}

We can define an equivalent ``do-calculus" like version of the causal effect definitions above for the case when a causal model defined as a graph structure is manipulated by an intervention that deletes an edge, or does some more sophisticated type of ``category" surgery. 

\section{Contributions and Future Work}

In this paper, we explored a layered hierarchical architecture for universal causality, based on the framework of category theory. Among the unique aspects of our formulation are the following novel contributions: 

\begin{enumerate}

\item {\bf Simplicial objects for causal ``surgery":} We introduced the framework of simplicial objects as a generic way to implement causal ``surgery" on a model, which defines contravariant functors from $\Delta$, the category of ordinal numbers, into a causal model. Simplicial objects provides an elegant and general way of extracting parts of a compositional structure, and its associated lifting problems define when a partial fragment of a causal model can be ``put back" together into a complete model. Such ``extension" problems, which were first explained at the beginning of the paper in connection with universal arrows, formalizes the notion of causal identifiability n in our framework. 

    \item {\bf Universal Arrow:} We identified the process of ``implementing" causal interventions from one categorical  layer to the next with the definition of a universal arrow from objects in the co-domain category to the functor mapping the domain category to the co-domain category. It is the property of universal arrows that permits mapping causal interventions from the simplicial layer down to the causal category layer, and from the causal category layer into the category of elements. 

    \item {\bf Lifting Problem:} Associated with each pair of layers of the UCLA hierarchy is a lifting problem over a suitable category of elements, from simplicial category of elements, to a category of elements over a dataset, to a category of elements over a topological space. In general, the Grothendieck category of elements is a way to embed each object in a category into the category of all categories {\bf Cat}. This construction has many elegant properties, which deserves further exploration in a subsequent paper. 

    \item {\bf Homotopy colimits and Classifying Spaces:} We defined causal effect in terms of the classifying space associated with the nerve of a causal category, and with the homotopy colimit of the nerve of the category of elements. These structures have been extensively explored in the study of homotopy in category theory \cite{richter2020categories}, and there are many advanced techniques that can be brought to bear on this problem. 
\end{enumerate}

\subsection{Future Work} 

There are many directions for future work, and we briefly describe a few of them below. 

\begin{enumerate}

\item {\bf Simplicial Causal Information Fields:}  Although we described one particular instantiation of our UCLA framework, there are many other ways to define the various categorical layers that we could not include for reasons of space. In particular, we can easily adapt our approach to capture the work on causal information fields \cite{cif}, where causal DAG models are formalized using Witsenhausen's measure-theoretic formalization of causality \cite{witsenhausen:1975}. In particular, \citet{heymann:arxiv} show that Pearl's do-calculus can be generalized using causal information fields to include causal models with feedback and other enhancements. In this setting, instead of conditional probability tables defining each variable of a model, a measure-theoretic information field is used that is defined as a measurable function from the product $\sigma$-algebras of its parents in the model. Witsenhausen defined causality as a condition on a given information field model of variables with a defined information field structure, whereby a given sequence of variables could be determined such that each variable's information field was uniquely computable given the values of the variables that preceded it in the ordering. Using the simplicial objects framework, we can define Witsenhausen's causality condition as a contravariant functor from the ordinal numbers $[n]$ that defines the partial ordering he used in his theoretical characterization of causality. In addition, we can define a modified category of elements that combines the causal category model with the category of measurable spaces, and reformulate his framework (and that of causal information fields) in our approach. 

\item {\bf Causal Discovery from Conditional Independence Oracles:} We did not discuss the issue of causal discovery at length in this paper, but many approaches in the literature rely on conditional independence oracles \cite{DBLP:conf/nips/KocaogluSB17}. In a previous paper, we defined categoroids \cite{categoroids}, a category-theoretic formulation of universal conditional independence, which generalizes well-known axiomatizations of conditional independence, such as {\em separoids} \cite{DBLP:journals/jmlr/Dawid10}, {\em graphoids} \cite{pearl:bnets-book} and {\em imsets} \cite{studeny2010probabilistic}. Categoroids are formally defined as a {\em join} of two categories \cite{kerodon}, combining a subcategory specifying a causal model, and another subcategory specifying the conditional independence structure of the model. It is possible to combine categoroids with the UCLA hierarchy, as it is known that simplicial objects can be defined over the join of two categories \cite{richter2020categories}. The problem of causal discovery can then be rigorously formulated as a lifting problem as well, using a simplicial extension of categoroids, where the conditional independence oracle is defined as a solution to a lifting problem (which is able to answer questions of the form $(x \CI y | \ z)$ (is $x$ conditionally independent of $y$ given $z$?). 

 \item {\bf Grothendieck Topology:} It is possible to define an abstract Grothendieck topology ${\cal J}$  for any category, which leads to the concept of a {\em site} \cite{maclane:sheaves}.  In simple terms, for any object $c$ in ${\cal C}$,  a {\em sieve} $S$ is a family of morphisms, all with co-domain $c$ such that 

    \[ f \in S \rightarrow f \circ g \in S \]

    for any $g$ where the composition is defined. A Grothendieck topology ${\cal J}$ on category ${\cal C}$ then defines a sieve $J(c)$ for each object $c$ such that the following properties hold: (i) the maximal sieve $t_c = \{f | cod(f) = c \}$ is in $J(c)$. There is an additional stability condition and a transitive closure condition. An interesting problem for future work is to define causal inference over sheaves of a site, using the concept of Grothendieck topologies. Any causal intervention that, for instance, deletes an edge, would cause a change in the structure of sieves. 

    \item {\bf Limitations of causal DAG models:} One reason for pursuing this line of research is to expand the scope of causal models to beyond well-studied paradigms, such as Bayesian networks and causal DAGs. \citet{studeny2010probabilistic} astutely observes that even in a relatively toy problem of $4$ variables, there are as many as $18,000$ conditional independence structures, but that DAG models can barely capture more than a few hundred of them. This fact suggests that DAG models are severely limited, and that there is much to be gained by exploring more powerful formalisms for causal inference. In particular, \citet{studeny2010probabilistic} himself proposed integer-valued sets (imsets) as a much more powerful formalism for representing conditional independence structures. The application of imsets to causal inference has not been well-studied, and the approach of constructing categories over imsets, which we explored in a previous paper \cite{categoroids}, may be an interesting direction for future research. 

    \item {\bf Causal confounding:} A notable omission of our paper is the lack of any discussion of the issue of confounding, one of the basic challenges facing any causal experiment in the real world. In the literature, many approaches to confounding have been studied, from models of missing data in statistics \cite{rubin-book} to using marginalized DAG and hyperedge DAG models \cite{mdag,hedge}. Usually, the assumption made is that while confounders are not directly observable, their impact on the underlying model is limited, and their effects can be ascertained under various assumptions. To treat confounding properly in a genuine category-theoretic manner is a problem that is outside the scope of this paper, and we have chosen to leave a detailed study of this important topic for future work. 
\end{enumerate}

\section{Summary} 

In this paper, we proposed a framework called Universal Causality (UC) for causal inference using the tools of category theory. Specifically, we described a layered hierarchical architecture called UCLA (Universal Causality Layered Architecture), where causal inference is modeled at multiple levels of categorical abstraction. At the top-most level, causal inference is modeled using a simplicial quasi-category of ordinal numbers $\Delta$, whose objects are the ordered natural numbers $[n] = \{0, \ldots, n\}$, and whose morphisms are order-preserving injections and surjections. Causal ``surgery" is then modeled as the action of a contravariant functor from the category $\Delta$ into a causal model.  At the second layer, causal models are defined by a category consisting of a collection of objects, such as the entities in a relational database, and morphisms between objects can be viewed  as attributes relating entities. The third categorical abstract layer corresponds to the data layer in causal inference, where each causal object is mapped into a set of instances, modeled using the category of sets and morphisms are functions between sets. The fourth layer comprises of additional structure imposed on the instance layer above, such as a topological space, a measurable space or a probability space, or more generally, a locale. Between every pair of layers in UCLA are functors that map objects and morphisms from the domain category to the co-domain category. Each functor between layers is characterized by a universal arrow, which defines an isomorphism between every pair of categorical layers. These universal arrows define universal elements and representations through the Yoneda Lemma, and in turn lead to a new category of elements based on a construction introduced by Grothendieck. Causal inference between each pair of layers is defined as a lifting problem, a commutative diagram whose objects are categories, and whose morphisms are functors that are characterized as different types of fibrations.  We defined causal effect in the UCLA framework using the notion of homotopy colimits associated with the nerve of a category.  We illustrate the UCLA architecture using a diverse set of examples. 

\newpage


\end{document}